\pdfoutput=1

\documentclass[11pt]{article}

\usepackage[final]{acl}
\usepackage{amsmath}
\usepackage{enumitem}
\allowdisplaybreaks[4]

\usepackage{times}
\usepackage{latexsym}

\usepackage[T1]{fontenc}

\usepackage[utf8]{inputenc}

\usepackage{microtype}

\usepackage{inconsolata}

\usepackage{graphicx}

\usepackage[utf8]{inputenc} 
\usepackage[T1]{fontenc}    
\usepackage{hyperref}       
\usepackage{url}            
\usepackage{booktabs}       
\usepackage{amsfonts}       
\usepackage{nicefrac}       
\usepackage{microtype}      
\usepackage{xcolor}         
\usepackage{graphicx}
\usepackage{multirow}
\usepackage{tablefootnote}
\usepackage{tcolorbox}
\usepackage{makecell}
\usepackage{CJKutf8}
\usepackage{subfigure}
\usepackage{bbding}

\newcommand{\hide}[1]{} 

\title{AlignMMBench: Evaluating Chinese Multimodal Alignment in Large Vision-Language Models}

\author{%
    Yuhang Wu$^{1,\ast,\dagger}$, Wenmeng Yu$^{2,\ast}$, Yean Cheng$^{3,\dagger}$, Yan Wang$^{2}$, \\
    \bf{Xiaohan Zhang$^{2}$, Jiazheng Xu$^{1,\dagger}$, Ming Ding$^{2}$, Yuxiao Dong$^{1}$} \\ \\
    $^1$Tsinghua University \hspace{0.3cm} $^2$Zhipu AI \hspace{0.3cm} $^3$Peking University
}

\begin{document}

\maketitle

\begin{figure*}[htbp]
\centering
\includegraphics[width=0.9\linewidth]{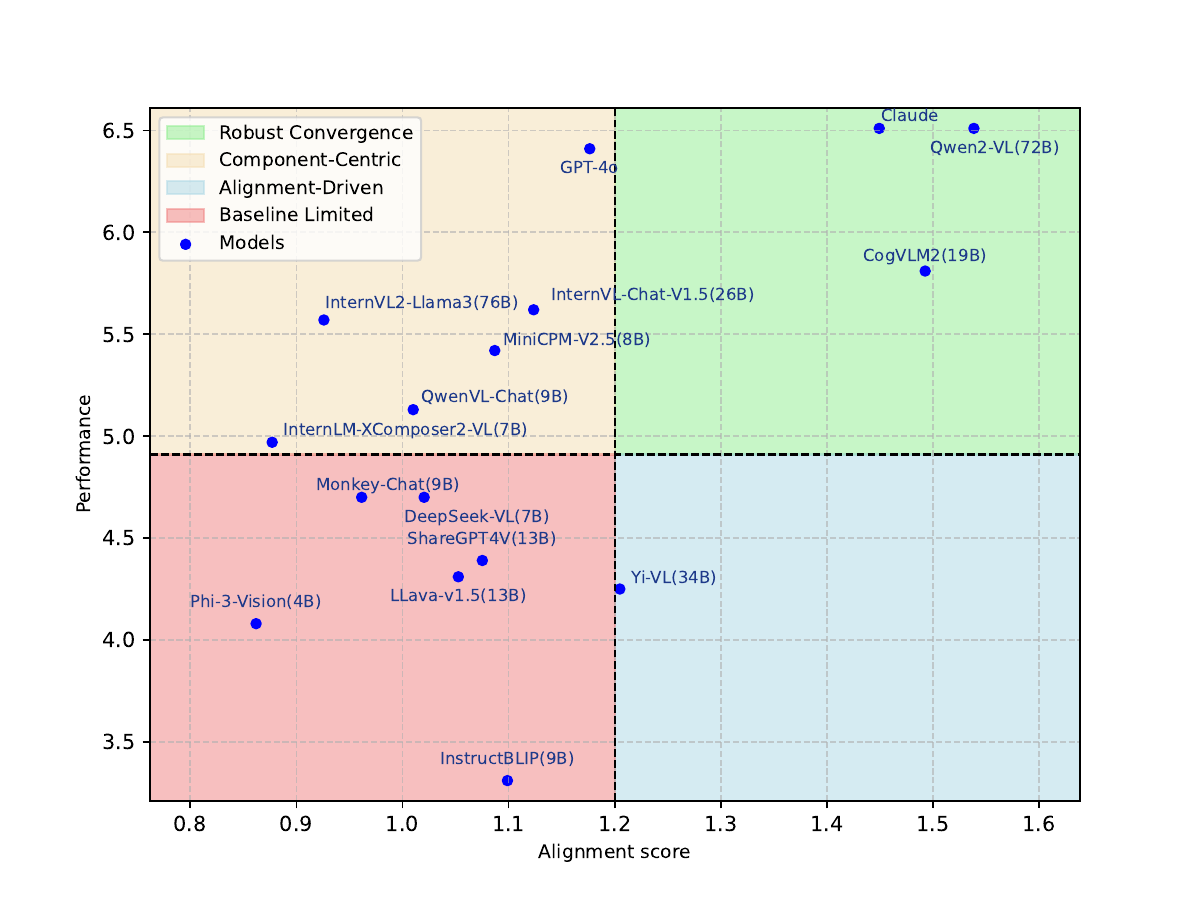}
\vspace{-2em}
\caption{Performance vs. Alignment score of various models. Performances are ranged from 0 to 10, while Alignment scores are ranged from 0.2 to $\infty$.}
\label{fig:comparison2d}
\vspace{-1em}
\end{figure*}

\renewcommand{\thefootnote}{\fnsymbol{footnote}}
    \footnotetext[1]{Equal contributions.}
    \footnotetext[2]{Work was partially done when interned at Zhipu AI.}
\renewcommand{\thefootnote}{\arabic{footnote}}

\begin{abstract} 
Evaluating the alignment capabilities of large Vision-Language Models (VLMs) is essential for determining their effectiveness as helpful assistants. 
However, existing benchmarks primarily focus on basic abilities using nonverbal methods, such as yes-no and multiple-choice questions. 
In this paper, we address this gap by introducing AlignMMBench, which provides more nuanced evaluations of alignment capabilities and is the first benchmark specifically designed for Chinese visual contexts. 
This benchmark is meticulously curated from real-world scenarios and internet sources, encompassing thirteen specific tasks across three categories, and includes both single-turn and multi-turn dialogue scenarios. 
Incorporating a prompt rewrite strategy, AlignMMBench encompasses 1,054 images and 4,978 question-answer pairs. 
To facilitate the evaluation pipeline, we develop CritiqueVLM, a rule-calibrated evaluator that exceeds GPT-4's evaluation ability. 
Additionally, we measure the “alignment score”, a quantitative metric designed to assess the robustness and stability of models across diverse prompts. 
Finally, we evaluate the performance of representative VLMs on AlignMMBench, offering insights into the capabilities and limitations of different VLM architectures. 
The evaluation code and data are available at \url{https://github.com/THUDM/AlignMMBench}.

\end{abstract}

\section{Introduction}
\label{introduction}

Equipped with Large Language Models (LLMs), Vision-Language Models (VLMs) demonstrate impressive performance in various visual tasks, such as image description and visual question answering~\cite{openai2024gpt4,wang2023cogvlm,Qwen-VL}. Following alignment training, like Supervised Fine-Tuning (SFT) and Reinforcement Learning from Human Feedback (RLHF), VLMs can comprehend and generate human language within visual contexts.

Recently, significant efforts have been made into developing VLMs for not only English but also Chinese contexts, 
such as QwenVL~\cite{Qwen-VL}, CogVLM~\cite{wang2023cogvlm}, MiniCPM~\cite{viscpm}, Intern-VL~\cite{chen2023internvl}, Intern-XComposer2V~\cite{internlmxcomposer2}, Deepseek-VL~\cite{lu2024deepseekvl}, and Yi-VL~\cite{ai2024yi}. 
Specifically, in publicly available comprehensive rankings\footnote{\scriptsize\url{https://rank.opencompass.org.cn/leaderboard-multimodal}}, these VLMs have demonstrated performance comparable to that of GPT-4o~\cite{openai2024gpt4}. To facilitate objective comparisons and evaluations among these models, researchers mainly utilize the following benchmarks: 1) general evaluation sets such as MME~\cite{fu2024mme} and MMBench~\cite{liu2023mmbench}, and 2) domain-specific evaluation sets such as MathVista~\cite{lu2023mathvista} and MMMU~\cite{yue2023mmmu}. However, these benchmarks primarily assess model capabilities through non-verbal methods and lack detailed evaluations of alignment performance, particularly in the Chinese context.

\begin{table*}[h]
\small
\centering
\setlength{\tabcolsep}{1.2mm}{
\begin{tabular}{l|ccccc|cc}
\toprule
\multirow{2}{*}{Benchmark} & \multicolumn{5}{c|}{Dataset}                                                                                                                                   & \multicolumn{2}{c}{Evaluation}                     \\ \cline{2-8} 
                           & \multicolumn{1}{c|}{Size} & \multicolumn{1}{c|}{Language}           & \multicolumn{1}{c|}{\makecell[c]{Multi\\Category}} & \multicolumn{1}{c|}{\makecell[c]{Dialogue \\Context}} & \makecell[c]{Open\\Ended} & \multicolumn{1}{c|}{\makecell[c]{Judge \\Method}} & Metric         \\ \hline
Ai2D~\cite{hiippala2021ai2d}          & \multicolumn{1}{c|}{4,903}  & \multicolumn{1}{c|}{English}            & \multicolumn{1}{c|}{\XSolidBrush}            & \multicolumn{1}{c|}{\XSolidBrush}               & \Checkmark        & \multicolumn{1}{c|}{multi-choice}          & Accuracy          \\
LLaVABench~\cite{liu2024visual}          & \multicolumn{1}{c|}{150}  & \multicolumn{1}{c|}{English}            & \multicolumn{1}{c|}{\Checkmark}            & \multicolumn{1}{c|}{\XSolidBrush}               & \Checkmark        & \multicolumn{1}{c|}{GPT}          & Score          \\
MathVista~\cite{lu2023mathvista}          & \multicolumn{1}{c|}{6,141}  & \multicolumn{1}{c|}{English\tablefootnote{MathVista contains 6.57\% non-English questions, such as Chinese and Persian.}}            & \multicolumn{1}{c|}{\XSolidBrush}            & \multicolumn{1}{c|}{\XSolidBrush}               & \Checkmark        & \multicolumn{1}{c|}{GPT}          & Accuracy          \\
MME~\cite{fu2024mme}       & \multicolumn{1}{c|}{2,800} & \multicolumn{1}{c|}{English}            & \multicolumn{1}{c|}{\Checkmark}            & \multicolumn{1}{c|}{\XSolidBrush}               & \XSolidBrush         & \multicolumn{1}{c|}{yes-no}      & Accuracy       \\
MMBench~\cite{liu2023mmbench}      & \multicolumn{1}{c|}{3,217} & \multicolumn{1}{c|}{Eng. \& Chi.} & \multicolumn{1}{c|}{\Checkmark}            & \multicolumn{1}{c|}{\XSolidBrush}               & \XSolidBrush         & \multicolumn{1}{c|}{multi-choice}          & Accuracy       \\
MMMU~\cite{yue2023mmmu}          & \multicolumn{1}{c|}{11,500}  & \multicolumn{1}{c|}{English}            & \multicolumn{1}{c|}{\XSolidBrush}            & \multicolumn{1}{c|}{\XSolidBrush}               & \XSolidBrush        & \multicolumn{1}{c|}{multi-choice}          & Accuracy          \\
MMStar~\cite{chen2024we}          & \multicolumn{1}{c|}{1,500}  & \multicolumn{1}{c|}{English}            & \multicolumn{1}{c|}{\Checkmark}            & \multicolumn{1}{c|}{\XSolidBrush}               & \XSolidBrush        & \multicolumn{1}{c|}{multi-choice}          & Accuracy          \\
MM-Vet~\cite{yu2023mmvet}          & \multicolumn{1}{c|}{205}  & \multicolumn{1}{c|}{English}            & \multicolumn{1}{c|}{\Checkmark}            & \multicolumn{1}{c|}{\XSolidBrush}               & \Checkmark        & \multicolumn{1}{c|}{GPT}          & Score          \\
OCRBench~\cite{liu2023hidden}          & \multicolumn{1}{c|}{1,000}  & \multicolumn{1}{c|}{English}            & \multicolumn{1}{c|}{\XSolidBrush}            & \multicolumn{1}{c|}{\XSolidBrush}               & \Checkmark        & \multicolumn{1}{c|}{string match}          & Accuracy          \\
TouchStone~\cite{bai2023touchstone}        & \multicolumn{1}{c|}{908}  & \multicolumn{1}{c|}{English}            & \multicolumn{1}{c|}{\Checkmark}            & \multicolumn{1}{c|}{\XSolidBrush}               & \Checkmark        & \multicolumn{1}{c|}{GPT}          & Score          \\
VisIT-Bench~\cite{bitton2023visitbench}      & \multicolumn{1}{c|}{592}  & \multicolumn{1}{c|}{English}            & \multicolumn{1}{c|}{\Checkmark}            & \multicolumn{1}{c|}{\XSolidBrush}               & \Checkmark        & \multicolumn{1}{c|}{GPT}          & Elo / Win \\
\textbf{AlignMMBench (ours)}       & \multicolumn{1}{c|}{\textbf{4,978}} & \multicolumn{1}{c|}{\textbf{Chinese}}            & \multicolumn{1}{c|}{\textbf{\Checkmark}}            & \multicolumn{1}{c|}{\textbf{\Checkmark}}              & \textbf{\Checkmark}        & \multicolumn{1}{c|}{\textbf{CritiqueVLM}}  & \textbf{Score}   \\       
\bottomrule

\end{tabular}
\footnotetext[1]{\small MathVista contains 6.57\% non-English questions, such as Chinese and Persian.}
}
\setlength{\abovecaptionskip}{0.1cm}
\caption{Comparisons between AlignMMBench and other benchmarks.}
\label{table:comparison}
\end{table*}

However, constructing a high-quality Chinese multimodal alignment benchmark is fully challenging. First, unlike the English context, which benefits from traditional benchmarks such as VQAv2~\cite{goyal2017making}, TextVQA~\cite{singh2019towards}, and ChartQA~\cite{masry2022chartqa}, clean and publicly available Chinese multimodal corpora are exceedingly scarce. It is necessary to collect and construct these resources from scratch. Second, the inherent ambiguity of the Chinese context is more pronounced~\cite{huang1997segmentation}, necessitating multiple annotators to repeatedly verify a single high-quality Chinese multimodal corpus. Besides, The characteristics of images and the underlying world knowledge can vary significantly between different languages~\cite{duncum2004visual}. As a result, relying solely on English datasets limits the comprehensive evaluation of Chinese vision-language models (VLMs). Therefore, developing a multimodal benchmark within Chinese visual context is essential and promising.

\begin{figure*}[!t]
\centering
\includegraphics[width=0.85\linewidth]{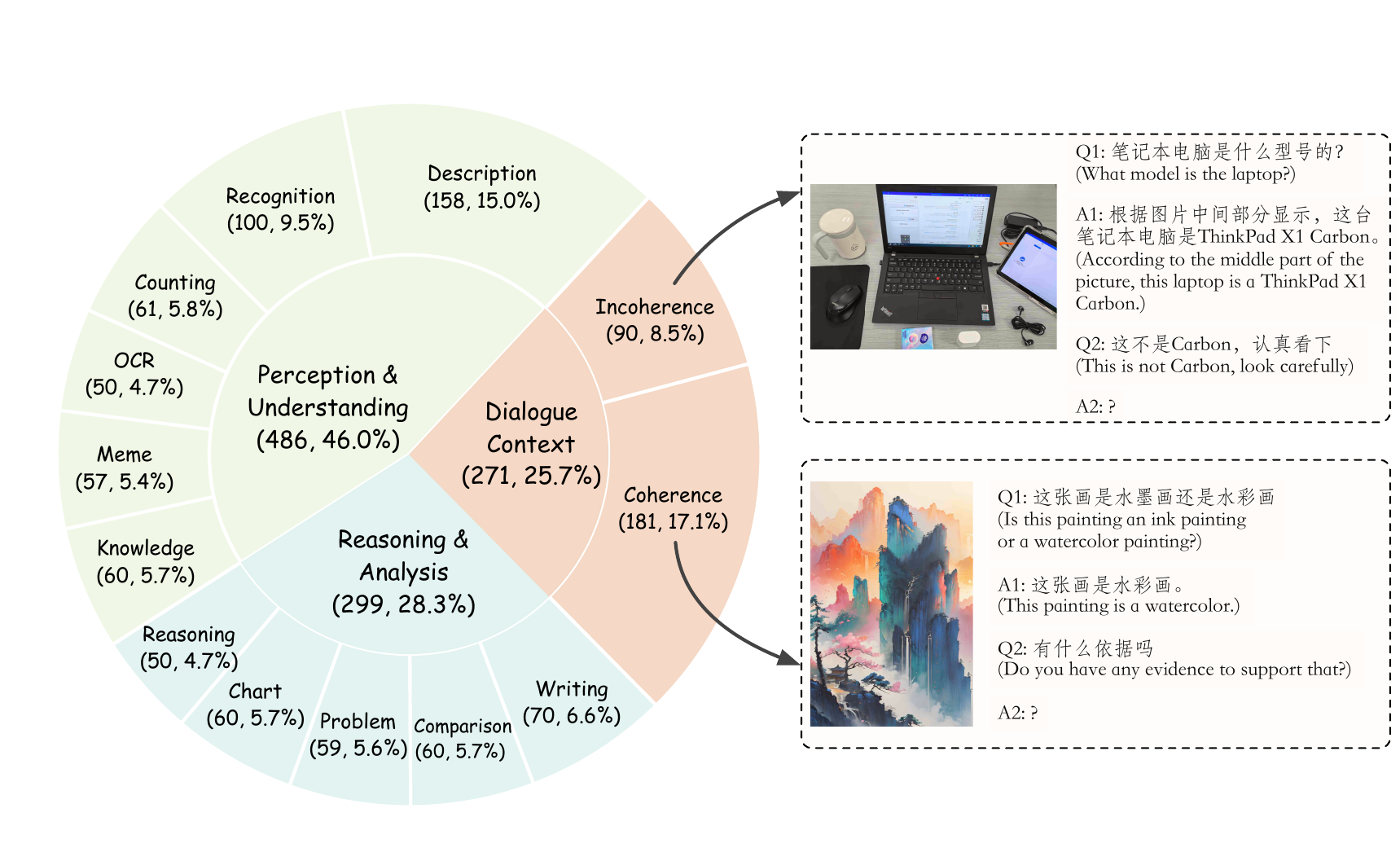}
\caption{Categories and examples of AlignMMBench. The chart on the left displays the categories of AlignMMBench, encompassing three main categories and thirteen specific tasks. The numbers listed under each category represent the number of images in that category and the corresponding percentage of the total. The right side of the pie chart presents two examples, illustrating instances from the incoherence and coherence tasks.}
\label{fig:PieChart}
\end{figure*}

To address these gap, we introduce AlignMMBench, a comprehensive evaluation benchmark specifically designed to assess the Chinese alignment capabilities. 
Given the scarcity of Chinese multimodal corpora, AlignMMBench is meticulously curated from real-world scenarios and internet resources.
It encompasses thirteen specific tasks across three high-level categories, including both single-turn and multi-turn dialogue scenarios, to make detailed evaluation of alignment performance.
Considering the diversity of query forms in real-world user scenarios, we introduce a LLM-based prompt-rewritting strategy, which transforms a single query into multiple stylistically distinct yet semantically equivalent questions. Leveraging this strategy, we introduce a new metric, the "alignment score", to investigate the reasons for performance differences among various models. By combining performance metrics with the alignment score, we categorize models into 4 classes, as specified in Figure \ref{fig:comparison2d}. As shown in Table \ref{table:comparison} and Figure \ref{fig:PieChart}, AlignMMBench comprises 1,054 images and 4,978 question-answer pairs.

Due to the lack of standard answers, the evaluation of open-ended questions is a significant and challenging research topic. One simple approach is to utilize the strongest LLM, such as GPT-4~\cite{openai2024gpt4}, to score the model responses. However, the API-based GPT-4 operates as a black box, resulting in limited control for developers.
The costs associated with API usage impose limitations on the size of the datasets that can be used as well as the number of models that can be tested.
To address these limitations, we fine-tuned an evaluator based on the ChatGLM3-6B~\cite{du2022glm}, named CritiqueVLM. Equipped with the detailed prompt rules and chain-of-thought reasoning, CritiqueVLM achieves a $34.8\%$ reduction in mean absolute error when compared to human scoring, outperforming GPT-4. 
Additionally, CritiqueVLM features higher consistency across other metrics.

Based on AlignMMBench and CritiqueVLM,we evaluate numerous popular VLMs that support Chinese. Our findings are as follows: (1) All VLMs excel in perception and understanding, achieving an average score of 5.07, but perform poorly in reasoning and analysis, with an average score of 4.38; (2) Within dialogue contexts, VLMs exhibit suboptimal performance in the incoherence task compared to the coherence task, indicating that VLMs struggle to detect previous errors; (3) Certain English-based VLMs, such as Phi-3-Vision~\cite{abdin2024phi}, exhibit suboptimal performance on AlignMMBench, suggesting the composition of the training corpus is critical in alignment evaluation.

The  contributions are summarized as follows:

\begin{itemize}[leftmargin=*, noitemsep, topsep=0pt]

\item We propose AlignMMBench, a multimodal alignment benchmark that encompasses both single-turn and multi-turn dialogue scenarios. It includes three categories and thirteen capability tasks, with a total of 4,978 question-answer pairs. As far as we know, this is the first public alignment benchmark specifically designed for the Chinese visual context.

\item To improve the controllability of evaluation scores, we introduce the CritiqueVLM, a ChatGLM3-6B based evaluator that has been rule-calibrated and carefully fine-tuned. With human judgements, its evaluation consistency surpasses that of GPT-4.

\item We benchmark representative VLMs on AlignMMBench. Beyond their performance, we provide in-depth insights into the current state of Chinese VLM and highlight the areas that require further enhancement.

\end{itemize}

\section{Related Work}
\label{related_work}

\paragraph{Multimodal benchmark.}

Early multimodal benchmarks predominantly focus on specific cross-modal tasks such as Image Caption~\cite{lin2014microsoft, plummer2015flickr30k}, Visual Grounding~\cite{kazemzadeh2014referitgame,yu2016modeling}, Visual Question Answering~\cite{schwenk2022okvqa, marino2019okvqa, mathew2021docvqa} and Optical Character Recognition (OCR)~\cite{singh2021textocr}. 
Recent benchmarks can be categorized into two types based on their questions, including domain-specific and general scenarios. 1) Domain-specific benchmarks concentrate on particular subfields. For example, discipline reasoning~\cite{yue2023mmmu, lu2023mathvista}, OCR~\cite{liu2023hidden}, chart understanding ~\cite{masry2022chartqa,hiippala2021ai2d}, and the hallucination test~\cite{liu2024visual}. 2) General benchmarks provide multi-dimensional and diverse assessments, encompassing MME~\cite{fu2024mme}, MMBench~\cite{liu2023mmbench}, MMStar~\cite{chen2024we}, MMVet~\cite{yu2023mmvet}, LLaVABench~\cite{chen2024we}, VisIT-Bench~\cite{bitton2023visitbench}, and TouchStone~\cite{bai2023touchstone}. 
From another perspective, based on differences in result evaluation methods, the aforementioned benchmarks can be divided into deterministic and open-ended types. 
Table \ref{table:comparison} provides an overview of the categories and salient features of prominent benchmarks.

Different from the above benchmarks, AlignMMBench is a universal and open-ended alignment benchmark. Additionally, we present dialogue context tasks based on real-world user corpus.

\paragraph{LLM-as-a-Judge methods.}

In the realm of automated model response evaluation, employing a robust LLM as an evaluation assistant is widely adopted across various VLM benchmarks. MathVista~\cite{lu2023mathvista} utilizes an LLM to extract answers from responses for more precise comparisons with reference answers, while other works~\cite{liu2024visual, yu2023mmvet, bai2023touchstone, bitton2023visitbench} adopt the LLM-as-a-Judge~\cite{zheng2024judging} approach. This method prompts the LLM to act as a judge, scoring responses or selecting the better response between two candidates. Furthermore, recent researches~\cite{ge2024mllmbenchevaluatingmultimodalllms, zhang2023gpt4visiongeneralistevaluatorvisionlanguage,kim2024prometheus2opensource,kim2023prometheus} explore the feasibility of employing the VLM-as-a-Judge approach to achieve more accurate evaluation results.

However, evaluating numerous models within a large benchmark can be costly and inefficient, with results potentially becoming unstable due to API-based model updates. The current capabilities of VLMs in instruction-following and logical reasoning are markedly inferior to those of the most advanced LLMs. In this work, we design a set of task-level evaluation prompts to improve the accuracy of LLM scoring. Additionally, we fine-tune ChatGLM3-6B~\cite{du2022glm} using a high-quality, human-annotated grading corpus to enhance the evaluation capabilities of our model.

\section{AlignMMBench}
\label{alignmmbench}


\subsection{Dataset Composition}

\begin{figure*}[!t]
\centering
\includegraphics[width=0.95\linewidth]{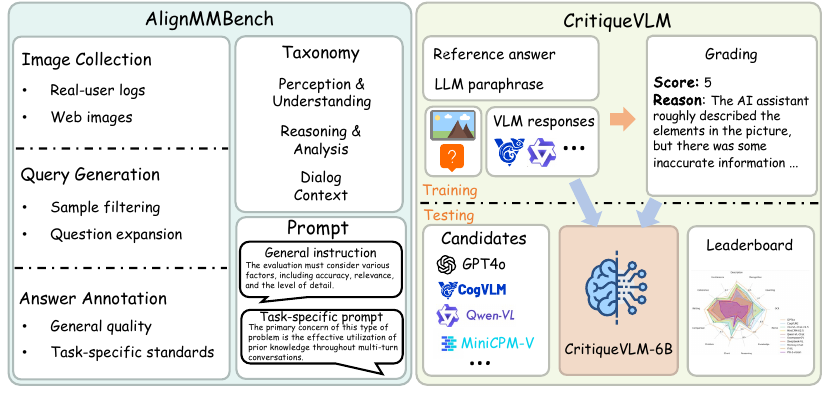}
\caption{Overall framework of our work.}
\label{fig:overview}
\end{figure*}

AlignMMBench comprises 1,056 images and 4,978 associated test cases. Each test case includes an image, a question, and a meticulously crafted reference answer. To evaluate the capabilities of VLMs across various dimensions, we categorized these test cases into three primary categories and thirteen distinct tasks, ranging from simple object recognition and description to complex reasoning and computation, as shown in Figure~\ref{fig:PieChart}. AlignMMBench consists of three major categories: \textbf{Perception and understanding}, which involves answering questions by synthesizing information from images and world knowledge; \textbf{Reasoning and analysis} focuses on assessing the model's capabilities in information processing and analysis, which often require numerical calculations or logical reasoning to provide accurate responses; \textbf{Dialogue context} evaluates capability in real-world user scenarios. 

For a detailed definition on these categories and their sub-tasks, please refer to Appendix \ref{benchdetail}.

\subsection{Dataset Construction}

As illustrated Figure~\ref{fig:overview} (left), the construction of AlignMMBench involves three steps: image collection, query generation, and answer annotation. 

\paragraph{Image collection.} 
First, we manually define 13 task types in AlignMMBench, with researchers constructing detailed descriptions and instructions for each type. Subsequently, web crawlers were employed to retrieve images from Chinese websites, such as Baidu, based on these predefined descriptions. To ensure data quality, researchers manually filtered out low-quality or irrelevant images from the dataset. Then, we collected real-world user queries from a Chinese application that offers generative AI assistant services, primarily catering to professionals and students. Throughout our process, we were diligent in avoiding the extraction of images from websites with copyright restrictions, and we conducted manual inspections to ensure adherence to copyright compliance.

Furthermore, we manually eliminate low-quality images based on the following protocols: (1) Removing images containing unidentifiable objects or text; (2) Eliminating images that contains personal privacy or offensive content; and (3) Excluding images with similar content that have already been included. Besides, we calculated the MD5 value of each image to avoid inclusion in our known SFT datasets or prior benchmarks.
\paragraph{Query generation.} For images obtained via the web crawler, we craft a seed question that correlates with the image and aligns with its designated category. Given the complexity of constructing a dialogue history based on an image, we use real-world user requests exclusively for building the "Dialogue Context" category. Considering the variability of problems with the same user intention, we employ ChatGLM~\cite{du2022glm} to rephrase the seed questions without altering their original purpose. This method was applied specifically to single-turn questions.
\paragraph{Answer annotation.} Annotators are instructed to produce accurate and comprehensive answers, incorporating as many pertinent details as possible. This approach enhances the ability to conduct precise evaluations independently of visual content.

Since we employ a prompt rewrite strategy to expand our question set, we implement a two-phase check process.
In the first phase, each seed questions-answer pair (1,054 pairs in total) is manually reviewed. Researchers conduct a thorough evaluation of the question-answer pairs, focusing on the following key aspects: 1) the relevance of the question to the image and the clarity of the answer, 2) the accuracy of the answer in addressing the question, and 3) the extent to which the answer provides sufficient detail for the independent reconstruction of the necessary information to address the question.
Subsequently, we utilize an LLM to expand the questions (details in Appendix \ref{augdetail}) and manually verify that each generated question is stylistically different yet semantically equivalent to seed questions. Following these checks, we obtained a total of 4,978 question-answer pairs.

Then, we manually eliminate low-quality responses based on the following error types: (1) \textbf{Factual errors:} The reference answer includes a description that does not correspond to the image content; (2) \textbf{Unfollowing:} The reference answer does not follow with the instruction in the prompt; (3) \textbf{Incomplete answers:} The reference answer lacks sufficient information to evaluate the response without the image; (4) \textbf{Common sense mistakes:} The reference answer contains common sense errors.

Finally, we employed a two-phase annotation process to obtain our responses. In the first turn, annotators craft responses based on the corresponding images. After that, researchers review these preliminary annotations, filter out low-quality entries, and request re-annotations where necessary.

These steps ensure a robust foundation for the subsequent phases of our study, providing a reliable and high-quality multimodal benchmark.
\begin{figure*}[!t]
\centering
\includegraphics[width=0.98\linewidth]{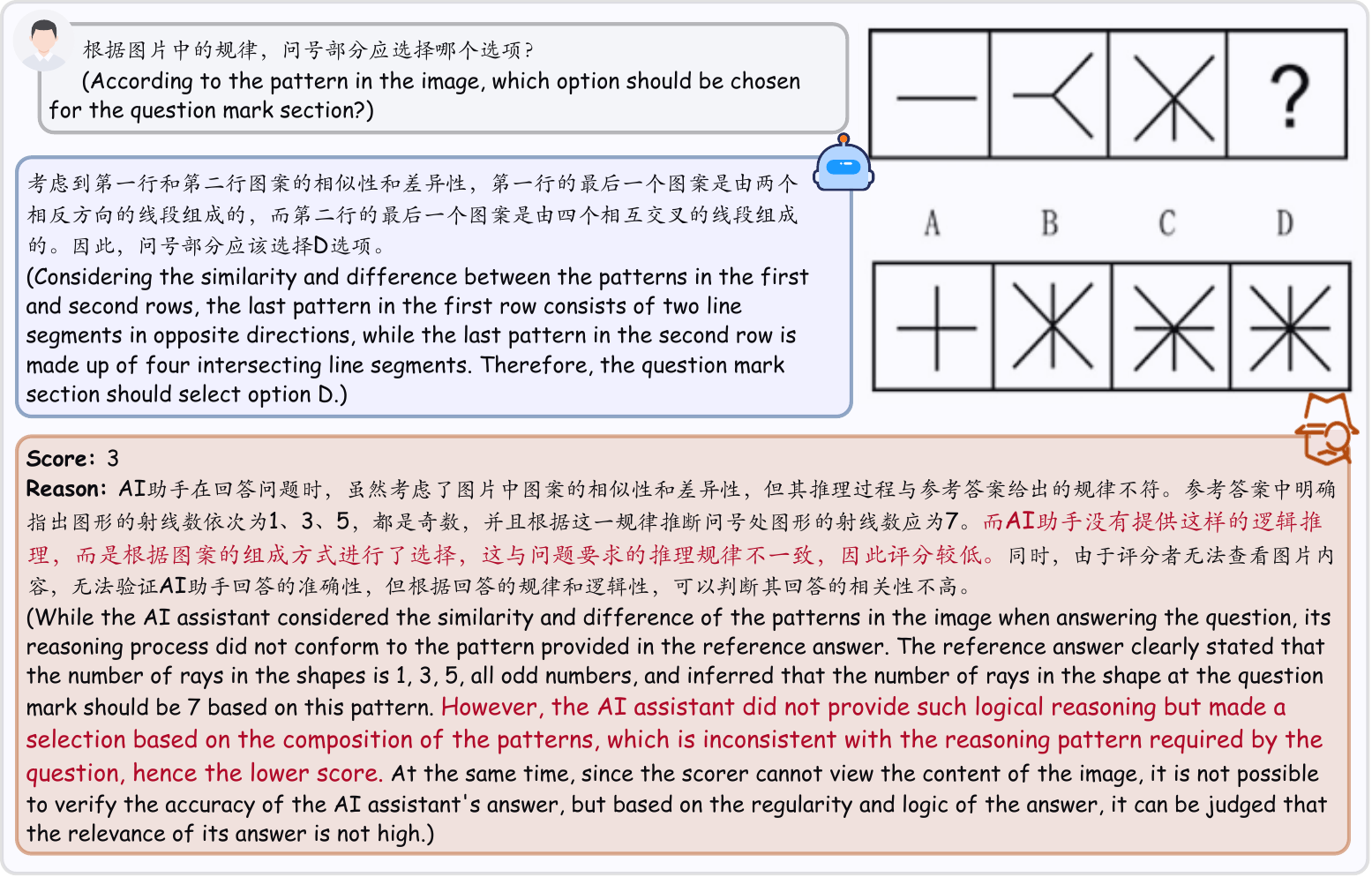}
\caption{A critical example from CritiqueVLM. The grey bubble is a query, the blue bubble is the response of the model being evaluated, and the red bubble is the output of the evaluator.}
\label{fig:example}
\end{figure*}

\section{CritiqueVLM}

In this section, we propose an LLM-based evaluator, referred to as CritiqueVLM, to automatically assess responses based on the provided questions and detailed reference answers. CritiqueVLM is fine-tuned from ChatGLM3-6B~\cite{du2022glm, zeng2022glm} and generates a score between 1 and 10, along with a chain-of-thought explanation, as illustrated in Figure~\ref{fig:example}.

\subsection{Training Data Construction}
\label{traiconstruct}
\paragraph{Prompt design.} Following ~\cite{zheng2024judging}, we devise a prompt system consisting of general and task-specific prompts. The general prompt outlines the scoring range, criteria, and output format. Task-specific prompts are varied by different categories, containing several hints to guide the critic model in accurately checking answers. Besides, we use in-context examples to mention that the critic model cannot access an image, encouraging it to infer the image's content. The detailed prompts are presented in Appendix \ref{evalprompt}.
\paragraph{Scoring data collection.} We collect the training corpus based on CogVLM~\cite{wang2023cogvlm} and QwenVL-Chat~\cite{Qwen-VL}. Given the challenging nature of the test cases in our AlignMMBench even for human annotators, we utilize responses from a "reference model" that rephrased reference answers using GLM-4~\cite{du2022glm, zeng2022glm}. This approach mitigates bias from predominantly low scores. Human annotators were then tasked with scoring each response from 1 to 10 and providing reasonable explanations, guided by the same prompts used by CritiqueVLM, but with access to images, questions, and reference answers. To prevent bias from the models' capabilities, we anonymize the model names and randomly shuffled the data.
\paragraph{Dataset partition.} We allocate 90\% of the data from each category to the training dataset and reserve the remaining 10\% as an internal testing dataset. To assess the model's agreement with humans when analyzing responses from a novel VLM, we create an external testing dataset using responses generated by GPT-4~\cite{openai2024gpt4} and annotated scores provided by human experts.

\subsection{Supervised Fine-Tuning} 
\label{sft}
We select ChatGLM3-6B~\cite{du2022glm, zeng2022glm} as the base model and fine-tune it on the training dataset. The DeepSpeed~\cite{rasley2020deepspeed} library was employed as our training framework, utilizing the FusedEmaAdam optimizer with a weight decay of 0.05. The learning rate is initially set to \(10^{-5}\) with a 1\% warmup ratio. The maximum source sequence length is set to 4096, and the target sequence length to 1024. Training is conducted with a global batch size of 128 on 32 A800 GPUs. After 1,000 iterations, the training loss decreased from 3.8 to 0.3.
\label{cogcritic}
\section{Experiments}
\label{experiments}

\begin{table*}[]
\centering 
\small
\renewcommand\arraystretch{1.2}
\setlength{\tabcolsep}{1.2mm}{
\begin{tabular}{l|cccccc|cccccc}
\toprule
Testcase        & \multicolumn{6}{c|}{Internal}                                                                                                                            & \multicolumn{6}{c}{External}                                                                                                                             \\ \midrule
Metric          & \multicolumn{1}{c|}{$e$}   & \multicolumn{1}{c|}{$r$}   & \multicolumn{1}{c|}{$\rho$} & \multicolumn{1}{c|}{$\tau$} & \multicolumn{1}{c|}{$f$}   & $s$   & \multicolumn{1}{c|}{$e$}   & \multicolumn{1}{c|}{$r$}   & \multicolumn{1}{c|}{$\rho$} & \multicolumn{1}{c|}{$\tau$} & \multicolumn{1}{c|}{$f$}   & $s$   \\ \midrule
ChatGLM3-6B      & \multicolumn{1}{c|}{2.424} & \multicolumn{1}{c|}{0.230} & \multicolumn{1}{c|}{0.224}  & \multicolumn{1}{c|}{0.194}  & \multicolumn{1}{c|}{0.350} & 0.285 & \multicolumn{1}{c|}{3.237} & \multicolumn{1}{c|}{0.103} & \multicolumn{1}{c|}{0.091}  & \multicolumn{1}{c|}{0.077}  & \multicolumn{1}{c|}{0.296} & 0.197 \\
ChatGPT         & \multicolumn{1}{c|}{1.720} & \multicolumn{1}{c|}{0.572} & \multicolumn{1}{c|}{0.596}  & \multicolumn{1}{c|}{0.505}  & \multicolumn{1}{c|}{0.427} & 0.347 & \multicolumn{1}{c|}{2.473} & \multicolumn{1}{c|}{0.404} & \multicolumn{1}{c|}{0.429}  & \multicolumn{1}{c|}{0.356}  & \multicolumn{1}{c|}{0.370} & 0.247 \\
GPT-4           & \multicolumn{1}{c|}{1.256} & \multicolumn{1}{c|}{0.839} & \multicolumn{1}{c|}{0.836}  & \multicolumn{1}{c|}{0.726}  & \multicolumn{1}{c|}{0.677} & 0.565 & \multicolumn{1}{c|}{1.486} & \multicolumn{1}{c|}{0.770} & \multicolumn{1}{c|}{0.765}  & \multicolumn{1}{c|}{0.648}  & \multicolumn{1}{c|}{0.550} & 0.424 \\
CritiqueVLM*  & \multicolumn{1}{c|}{\textbf{0.818}} & \multicolumn{1}{c|}{\textbf{0.846}} & \multicolumn{1}{c|}{\textbf{0.838}}  & \multicolumn{1}{c|}{\textbf{0.740}}  & \multicolumn{1}{c|}{\textbf{0.747}} & \textbf{0.646} & \multicolumn{1}{c|}{\textbf{1.146}} & \multicolumn{1}{c|}{\textbf{0.778}} & \multicolumn{1}{c|}{\textbf{0.782}}  & \multicolumn{1}{c|}{\textbf{0.671}}  & \multicolumn{1}{c|}{\textbf{0.670}} & \textbf{0.511} \\
\bottomrule
\end{tabular}}
\setlength{\abovecaptionskip}{0.1cm}
\caption{Results of the agreement between human annotated scores and models. The best scores are in \textbf{bold}.}
\label{mainresulttable}
\end{table*}


\subsection{Baselines}
We select two models, {ChatGPT} (gpt-3.5-turbo) and {GPT-4} (gpt-4-1106-preview)\cite{openai2024gpt4}, as baselines for CritiqueVLM, as they are widely utilized in other benchmark studies. Additionally, given that AlignMMBench is a Chinese-style benchmark, {ChatGLM3-6B}\cite{du2022glm,zeng2022glm} is also included. These baselines are evaluated in a zero-shot setting.

\subsection{Evaluation of Evaluator Performance}
\label{mainres}

In our experiment, we employ six statistical metrics to assess the agreement between human annotated scores and model generated scores. We use the following four common statistical metrics: the mean absolute error ($e$), Pearson ($r$), Spearman ($\rho$), and Kendall correlation coefficient ($\tau$). To mitigate the bias from annotators preferences, we map scores to predefined ranges and calculate accuracy as an evaluation metric. Based on scoring criteria, integers from 1 to 10 are divided into two sets of ranges:

\begin{itemize}[leftmargin=*, noitemsep, topsep=0pt]
    \item \textbf{Fuzzy division.} This division includes 4 ranges: Unfollow ($[1,2]$), Bad ($[3,5]$), Good ($[6,8]$), and Excellent ($[9,10]$).
    We consider a model's score to be valid if and only if it falls within the same range as the human score. The final evaluation score ($f$) of the model is defined as the proportion of cases where this criterion is met.
    \item \textbf{Strict division.} This division includes 7 ranges, each corresponding to specific ranges defined in the scoring criteria: $[1,1],[2,2],[3,3],[4,5],[6,6],[7,8],[9,10]$.
    The specific method for computing the model score ($s$) is the same as described above.
    
\end{itemize}

Detailed results are presented in Table~\ref{mainresulttable}, demonstrating that CritiqueVLM achieves superior performance with only 6 billion parameters. Additionally, Figure~\ref{fig:example} provides two examples of scoring by CritiqueVLM, validating its ability to generate reasonable critical scores and detailed explanations.
\subsection{Leaderboard}
\label{leaderboard}

\begin{table*}[h]
\centering
\small
\renewcommand\arraystretch{1.25}
\setlength{\tabcolsep}{0.8mm}{ 
\begin{tabular}{l|c|c|c|cccccc|ccccc|cc|c}
\toprule
 \multirow{3}{*}{Models} & 
 \multirow{3}{*}{Size} &
 \multirow{3}{*}{Ref.} &
 \multirow{3}{*}{Avg}  & 
 \multicolumn{6}{c|}{Perception \& Understanding.} & 
 \multicolumn{5}{c|}{Reasoning \& Analysis} & 
 \multicolumn{2}{c|}{Context} &
 \multirow{3}{*}{Align.} \\
\cline{5-17}
    & & & & Des. & Rec. & Cou. & OCR. & Mem. & \multicolumn{1}{c|}{Kno.} & Rea. & Cha. & Pro. & Com. & \multicolumn{1}{c|}{Wri.}  & Coh. & Inc. &  \\
\midrule
Qwen2-VL & 72B & 1 & \textbf{6.51} &  7.39 & \underline{6.64} & \underline{6.64} & \textbf{7.60} & 7.09 & \multicolumn{1}{c|}{\textbf{6.32}} & 4.00 & \underline{7.16} & \textbf{5.89} & \underline{6.57} & \multicolumn{1}{c|}{7.72} & 6.37 & \underline{5.26} & \textbf{1.54} \\
Claude & - & 4 & \textbf{6.51} &  \underline{7.68} & \textbf{6.89} & \textbf{6.79} & 7.02 & \underline{7.10} & \multicolumn{1}{c|}{\underline{6.28}} & \underline{4.06} & 7.11 & 5.20 & 5.92 & \multicolumn{1}{c|}{\textbf{7.98}} & \textbf{7.02} & \textbf{5.52} & 1.45 \\
GPT-4o & - & 2 & \underline{6.41} &  \textbf{7.75} & 6.41 & 5.20 & 7.17 & \textbf{7.28} & \multicolumn{1}{c|}{6.16} & \textbf{4.44} & \textbf{7.23} & \underline{5.81} & \textbf{7.19} & \multicolumn{1}{c|}{\underline{7.85}} & \underline{6.41} & 4.43 & 1.18 \\
CogVLM2& 19B & 8 &5.81 & 7.20 & 6.12 & 5.75 & 7.21 & 6.07 & \multicolumn{1}{c|}{5.69}  & 3.43 & 5.92 & 4.37 & 5.65 & \multicolumn{1}{c|}{7.34} & 6.33 & 4.43 & \underline{1.49} \\
InternVL-Chat & 26B & 5 & 5.62 & 7.12 & 6.00 & 5.51 & 6.63 & 4.99 & \multicolumn{1}{c|}{5.08} & 3.35 & 5.98 & 3.98 & 6.33 & \multicolumn{1}{c|}{7.26} & 6.31 & 4.48 & 1.12 \\
InternVL2 & 76B & 3 & 5.57 & 6.95 & 5.11 & 5.81 & \underline{7.37} & 5.96 & 3.61 & 3.83 & 6.48 & 4.66 & 6.05 & 6.05 & 6.30 & 4.23 & 0.93 \\
MiniCPM& 8B & 6 & 5.42  & 7.18 & 5.37 & 5.46 & 6.23 & 4.46 & \multicolumn{1}{c|}{5.35} & 3.34 & 4.83 & 3.69 & 5.99 & \multicolumn{1}{c|}{7.35} & 6.25 & 4.97 & 1.09 \\
Qwen-VL-Chat & 9B & 12 & 5.13 & 6.43 & 5.87 & 5.40 & 4.80 & 5.11 & \multicolumn{1}{c|}{5.58} & 2.98 & 4.10 & 3.12 & 5.51 & \multicolumn{1}{c|}{7.19} & 6.07 & 4.50& 1.01 \\
InternLM-XC2-VL & 7B & 7 & 4.97 & 6.34 & 4.70 & 5.28 & 5.06 & 4.69 & \multicolumn{1}{c|}{5.03} & 3.08 & 4.49 & 3.29 & 5.00 & \multicolumn{1}{c|}{7.21} & 5.92 & 4.56& 0.88  \\
DeepSeek-VL & 7B & 11 & 4.70 & 6.53 & 5.52 & 5.10 & 3.98 & 3.87 & \multicolumn{1}{c|}{4.19} & 2.50 & 3.96 & 2.58 & 5.46 & \multicolumn{1}{c|}{7.15} & 5.83 & 4.47 & 1.02 \\
Monkey-Chat & 9B & 10 & 4.70 & 6.04 & 4.88 & 5.57 & 4.66 & 4.18 & \multicolumn{1}{c|}{4.96} & 3.01 & 4.00 & 2.61 & 4.87 & \multicolumn{1}{c|}{6.29} & 6.15 & 3.96 & 0.96 \\
ShareGPT4V & 13B & 14 & 4.39 & 5.93 & 4.61 & 5.16 & 3.77 & 4.04 & \multicolumn{1}{c|}{4.58} & 2.45 & 3.73 & 2.19 & 5.05 & \multicolumn{1}{c|}{6.39} & 5.36 & 3.79 & 1.08  \\
LLava-v1.5 & 13B & 15 & 4.31 & 6.02 & 4.56 & 4.46 & 3.85 & 3.69 & \multicolumn{1}{c|}{4.72} & 2.46 & 3.69 & 2.10 & 4.75 & \multicolumn{1}{c|}{6.21} & 5.60 & 3.96 & 1.05 \\
Yi-VL & 34B & 13 & 4.25 & 4.79 & 4.78 & 5.19 & 3.33 & 3.58 & \multicolumn{1}{c|}{4.47} & 2.42 & 3.25 & 2.08 & 4.72 & \multicolumn{1}{c|}{6.61} & 5.87 & 4.13 & 1.20 \\
Phi-3-Vision & 4B & 9 & 4.08 & 4.48 & 3.53 & 4.75 & 4.10 & 3.48 & \multicolumn{1}{c|}{3.16} & 2.56 & 4.40 & 2.85 & 4.34 & \multicolumn{1}{c|}{5.51} & 5.85 & 4.07 & 0.86 \\
InstructBLIP & 9B & 16 & 3.31 & 4.11 & 4.61 & 4.11 & 2.77 & 3.05 & \multicolumn{1}{c|}{2.92} & 1.76 & 2.58 & 1.12 & 3.36 & \multicolumn{1}{c|}{3.17} & 5.42 & 4.02 & 1.09 \\
\midrule 
GPT-4o without image & - & - & 2.13 &  1.11 & 1.57 & 1.22 & 1.73 & 1.53 & \multicolumn{1}{c|}{1.17} & 1.29 & 2.88 & 1.14 & 1.99 & \multicolumn{1}{c|}{3.50} & 5.14 & 3.41 & -  \\
\bottomrule
\end{tabular}}
\setlength{\abovecaptionskip}{0.1cm}
\caption{Evaluation results on AlignMMBench.
For each column, the highest score is \textbf{bold}, while the second highest score is \underline{underlined}. The "Ref." column indicates the relative ranking of these models on the \url{https://rank.opencompass.org.cn/leaderboard-multimodal}, dominated by primarily English benchmarks. Table \ref{summarytable} presents the detailed versions and architectures of these open-sourced models.}
\label{table:leaderboard}
\end{table*}

We benchmark a range of popular VLMs, including GPT4o~\cite{gpt4o}, Qwen2-VL~\cite{Qwen2VL}, CogVLM2~\cite{wang2023cogvlm}, InternVL2, InternVL-Chat-V1.5~\cite{chen2023internvl, chen2024far, chen2024expanding, gao2024mini}, MiniCPM-V2.5~\cite{yu2023rlhf, viscpm, xu2024llava-uhd, yu2024rlaifv}, Qwen-VL-Chat~\cite{Qwen-VL}, XComposer2V~\cite{internlmxcomposer2}, DeepSeek-VL~\cite{lu2024deepseekvl}, Monkey-Chat~\cite{li2023monkey}, Yi-VL~\cite{ai2024yi}, Phi-3-Vision~\cite{abdin2024phi}, ShareGPT4V~\cite{chen2023sharegpt4v}, LLava-v1.5~\cite{liu2024visual}, and InstructBLIP~\cite{instructblip}. Results are shown in Table \ref{mainresulttable}.

\subsubsection{Analysis of task performance} 

The average scores of VLMs in Table~\ref{table:leaderboard} range from 3.3 to 6.5, indicating that most VLMs can understand question requirements and generate responses relevant to the images, according to our scoring criteria. Scores below 5.0 reflect numerous errors in the VLMs' responses. Additionally, GPT4o~\cite{gpt4o} demonstrates the best performance across most tasks, and CogVLM2~\cite{wang2023cogvlm} secures the second-best performance. 

Moreover, the disparity in rankings between the "Ref." column and AlignMMBench underscores the limitations of existing benchmarks, which do not fully account for characteristics in Chinese context. Consequently, AlignMMBench serves as a valuable complement to existing benchmarks.


\subsubsection{Analysis of alignment ability} 

We propose a novel metric, "alignment score", to evaluate the alignment capabilities of VLMs. Consider a dataset with $N$ seed questions, where each seed question, denoted as $S_i$, generates a set of $M_i$ semantically equivalent questions $\{Q_{i,1},Q_{i,2}\ldots,Q_{i,M_i}\}$. For each question $Q_{i,j}$, let $R_{i,j}$ represent the score achieved by the model on this question. The proposed metric is defined as the inverse of the average standard deviation across all seed questions and is expressed as follows:

{\small
\begin{align*}
\text{Align.} &= \frac{N}{\sum_{i=1}^N \sigma_i}\\
\sigma_i &= \sqrt{\frac{1}{M_i}\sum_{j=1}^{M_i}(R_{i,j}-\overline{R}_i)^2}\\
\overline{R}_i &= \frac{1}{M_i}\sum_{j=1}^{M_i} R_{i,j}
\end{align*}
}

This metric reflects the average variability of results within clusters of semantically equivalent questions, with higher values indicating greater consistency.
We agree that a well-aligned model can demonstrate consistent performance when presented with stylistically distinct yet
semantically equivalent questions. Accordingly, this metric is selected as an evaluation criterion to assess the model's alignment capabilities.

By combining the average score and the alignment score, these models can be categorized into four groups, as illustrated in Figure \ref{fig:comparison2d}:

\begin{itemize}[leftmargin=*, noitemsep, topsep=0pt]
    \item \textbf{Robust Convergence}: These models exhibit a strong capacity for alignment, with both their visual and language components demonstrating robustness to achieve high levels of performance.
    \item \textbf{Component-Centric}: While the visual and language parts in these models exhibit notable strengths, further advancements are necessary to enhance their multimodal alignment.
    \item \textbf{Alignment-Driven}: These models display effective multimodal alignment; however, their overall performance remains constrained by inherent limitations in their visual or language parts.
    \item \textbf{Baseline Limited}: These models exhibit limited capabilities in both alignment and their visual/language parts, necessitating further improvements to enhance performance.
\end{itemize}


\subsubsection{Analysis on category-level performance}

\paragraph{Single-turn scenarios.}

Focusing on VLMs with average scores above 6.0, we observe that they perform well in tasks such as description, OCR, and writing. These tasks require VLMs to understand images but do not include complex reasoning or computation. However, they do not perform well in tasks involving reasoning and problem solving. For instance, GPT4o~\cite{gpt4o} scores only 4.44 in reasoning and 5.81 in problems, indicating frequent response errors. These observations suggest that while current top-tier VLMs can comprehend images and integrate information from images and texts, they struggle with test cases that demand complex reasoning and computation.

\paragraph{Multi-turn scenarios.} 

In the coherence task, GPT-4o~\cite{gpt4o} demonstrates the best performance among all models. Most VLMs achieve scores above 6.0, indicating their ability to follow instructions from previous interactions and utilize information from the dialogue context effectively. However, all VLMs exhibit suboptimal performance in the incoherence task. This suggests that these models struggle to detect previous errors within the dialogue context and to make accurate corrections as guided by users.

\subsubsection{Analysis on different backbone}
\label{appendix-analysis}
To better illustrate the importance of our benchmark in Chinese, we conducted a survey of the open-source models featured in our leaderboard. The results are presented in Table \ref{summarytable}. By integrating their underlying architectures with their performance metrics from the leaderboard, we make these observations:

\begin{itemize}
\item The training corpus plays an important role in model performance. As shown in the Table \ref{summarytable}, models primarily trained in Chinese consistently outperform their English counterparts on AlignMMBench. Consequently, it can be inferred that previous evaluations of Chinese multimodal models based on English benchmarks may not have been sufficient and satisfactory.

\item For both Chinese and English models, there is a generally positive correlation between model size and performance scores, indicating that larger and more recent models tend to achieve better results.

\item Due to the variation in training corpora across different models, it is challenging to assess the specific impact of different model architectures. This issue represents an important research topic that extends beyond the scope of this paper's discussion.
\end{itemize}
\section{Conclusion}
\label{conclusion}

In this paper, we introduce AlignMMBench, a comprehensive Chinese multimodal alignment benchmark comprising three high-level categories and thirteen subtasks. AlignMMBench includes 1,054 images and 4,978 question-answer pairs, encompassing both single-turn and multi-turn dialogue scenarios. To facilitate accurate and efficient evaluations, we developed a critique generation model, referred to as CritiqueVLM. Experimental results demonstrate that CritiqueVLM can assign scores aligned with human preferences and achieve superior performance compared to the widely used GPT-4. Additionally, we present a leaderboard featuring popular VLMs, highlighting potential directions for future improvements in VLMs. We anticipate that this dataset will further advance the development of multimodal language models.

\section*{Limitations}
\label{limit}

First, AlignMMBench functions as an alignment benchmark within Chinese context and does not evaluate the multilingual capabilities of VLMs. We plan to gather more images and questions in other languages to extend the scope of our benchmark. 
Second, due to the suboptimal performance of VLMs in scoring evaluation tasks~\cite{chen2024mllm}, we currently employ an LLM as our evaluator, which necessitates the use of reference answers that exclude image-derived information. In future work, we will explore the feasibility of using VLMs as reliable evaluators without requiring human-annotated reference answers.






\section*{Acknowledgment}
This work is supported by the Natural Science Foundation of China (NSFC) 62276148 and 62495063. 
The GPU compute used in this work is sponsored by Zhipu AI. 
The corresponding author: Yuxiao Dong (yuxiaod@tsinghua.edu.cn).

\clearpage
\bibliography{ref}

\begin{thebibliography}{54}
\providecommand{\natexlab}[1]{#1}

\bibitem[{Abdin et~al.(2024)Abdin, Jacobs, Awan, Aneja, Awadallah, Awadalla, Bach, Bahree, Bakhtiari, Behl et~al.}]{abdin2024phi}
Marah Abdin, Sam~Ade Jacobs, Ammar~Ahmad Awan, Jyoti Aneja, Ahmed Awadallah, Hany Awadalla, Nguyen Bach, Amit Bahree, Arash Bakhtiari, Harkirat Behl, et~al. 2024.
\newblock Phi-3 technical report: A highly capable language model locally on your phone.
\newblock \emph{arXiv preprint arXiv:2404.14219}.

\bibitem[{Achiam et~al.(2023)Achiam, Adler, Agarwal, Ahmad, Akkaya, Aleman, Almeida, Altenschmidt, Altman, Anadkat et~al.}]{openai2024gpt4}
Josh Achiam, Steven Adler, Sandhini Agarwal, Lama Ahmad, Ilge Akkaya, Florencia~Leoni Aleman, Diogo Almeida, Janko Altenschmidt, Sam Altman, Shyamal Anadkat, et~al. 2023.
\newblock Gpt-4 technical report.
\newblock \emph{arXiv preprint arXiv:2303.08774}.

\bibitem[{AI et~al.(2024)AI, :, Young, Chen, Li, Huang, Zhang, Zhang, Li, Zhu, Chen, Chang, Yu, Liu, Liu, Yue, Yang, Yang, Yu, Xie, Huang, Hu, Ren, Niu, Nie, Xu, Liu, Wang, Cai, Gu, Liu, and Dai}]{ai2024yi}
01. AI, :, Alex Young, Bei Chen, Chao Li, Chengen Huang, Ge~Zhang, Guanwei Zhang, Heng Li, Jiangcheng Zhu, Jianqun Chen, Jing Chang, Kaidong Yu, Peng Liu, Qiang Liu, Shawn Yue, Senbin Yang, Shiming Yang, Tao Yu, Wen Xie, Wenhao Huang, Xiaohui Hu, Xiaoyi Ren, Xinyao Niu, Pengcheng Nie, Yuchi Xu, Yudong Liu, Yue Wang, Yuxuan Cai, Zhenyu Gu, Zhiyuan Liu, and Zonghong Dai. 2024.
\newblock Yi: Open foundation models by 01.ai.
\newblock \emph{arXiv preprint arXiv:2403.04652}.

\bibitem[{Bai et~al.(2023{\natexlab{a}})Bai, Bai, Yang, Wang, Tan, Wang, Lin, Zhou, and Zhou}]{Qwen-VL}
Jinze Bai, Shuai Bai, Shusheng Yang, Shijie Wang, Sinan Tan, Peng Wang, Junyang Lin, Chang Zhou, and Jingren Zhou. 2023{\natexlab{a}}.
\newblock Qwen-vl: A versatile vision-language model for understanding, localization, text reading, and beyond.
\newblock \emph{arXiv preprint arXiv:2308.12966}.

\bibitem[{Bai et~al.(2023{\natexlab{b}})Bai, Yang, Bai, Wang, Zhang, Lin, Wang, Zhou, and Zhou}]{bai2023touchstone}
Shuai Bai, Shusheng Yang, Jinze Bai, Peng Wang, Xingxuan Zhang, Junyang Lin, Xinggang Wang, Chang Zhou, and Jingren Zhou. 2023{\natexlab{b}}.
\newblock Touchstone: Evaluating vision-language models by language models.
\newblock \emph{arXiv preprint arXiv:2308.16890}.

\bibitem[{Bitton et~al.(2023)Bitton, Bansal, Hessel, Shao, Zhu, Awadalla, Gardner, Taori, and Schmidt}]{bitton2023visitbench}
Yonatan Bitton, Hritik Bansal, Jack Hessel, Rulin Shao, Wanrong Zhu, Anas Awadalla, Josh Gardner, Rohan Taori, and Ludwig Schmidt. 2023.
\newblock Visit-bench: A benchmark for vision-language instruction following inspired by real-world use.
\newblock \emph{arXiv preprint arXiv:2308.06595}.

\bibitem[{Chen et~al.(2024{\natexlab{a}})Chen, Chen, Zhang, Liu, Wang, Zhou, Zhang, Zhou, Wan, and Sun}]{chen2024mllm}
Dongping Chen, Ruoxi Chen, Shilin Zhang, Yinuo Liu, Yaochen Wang, Huichi Zhou, Qihui Zhang, Pan Zhou, Yao Wan, and Lichao Sun. 2024{\natexlab{a}}.
\newblock Mllm-as-a-judge: Assessing multimodal llm-as-a-judge with vision-language benchmark.
\newblock \emph{arXiv preprint arXiv:2402.04788}.

\bibitem[{Chen et~al.(2024{\natexlab{b}})Chen, Li, Dong, Zhang, Zang, Chen, Duan, Wang, Qiao, Lin et~al.}]{chen2024we}
Lin Chen, Jinsong Li, Xiaoyi Dong, Pan Zhang, Yuhang Zang, Zehui Chen, Haodong Duan, Jiaqi Wang, Yu~Qiao, Dahua Lin, et~al. 2024{\natexlab{b}}.
\newblock Are we on the right way for evaluating large vision-language models?
\newblock \emph{arXiv preprint arXiv:2403.20330}.

\bibitem[{Chen et~al.(2023{\natexlab{a}})Chen, Li, Dong, Zhang, He, Wang, Zhao, and Lin}]{chen2023sharegpt4v}
Lin Chen, Jisong Li, Xiaoyi Dong, Pan Zhang, Conghui He, Jiaqi Wang, Feng Zhao, and Dahua Lin. 2023{\natexlab{a}}.
\newblock Sharegpt4v: Improving large multi-modal models with better captions.
\newblock \emph{arXiv preprint arXiv:2311.12793}.

\bibitem[{Chen et~al.(2024{\natexlab{c}})Chen, Wang, Cao, Liu, Gao, Cui, Zhu, Ye, Tian, Liu et~al.}]{chen2024expanding}
Zhe Chen, Weiyun Wang, Yue Cao, Yangzhou Liu, Zhangwei Gao, Erfei Cui, Jinguo Zhu, Shenglong Ye, Hao Tian, Zhaoyang Liu, et~al. 2024{\natexlab{c}}.
\newblock Expanding performance boundaries of open-source multimodal models with model, data, and test-time scaling.
\newblock \emph{arXiv preprint arXiv:2412.05271}.

\bibitem[{Chen et~al.(2024{\natexlab{d}})Chen, Wang, Tian, Ye, Gao, Cui, Tong, Hu, Luo, Ma et~al.}]{chen2024far}
Zhe Chen, Weiyun Wang, Hao Tian, Shenglong Ye, Zhangwei Gao, Erfei Cui, Wenwen Tong, Kongzhi Hu, Jiapeng Luo, Zheng Ma, et~al. 2024{\natexlab{d}}.
\newblock How far are we to gpt-4v? closing the gap to commercial multimodal models with open-source suites.
\newblock \emph{arXiv preprint arXiv:2404.16821}.

\bibitem[{Chen et~al.(2023{\natexlab{b}})Chen, Wu, Wang, Su, Chen, Xing, Zhong, Zhang, Zhu, Lu, Li, Luo, Lu, Qiao, and Dai}]{chen2023internvl}
Zhe Chen, Jiannan Wu, Wenhai Wang, Weijie Su, Guo Chen, Sen Xing, Muyan Zhong, Qinglong Zhang, Xizhou Zhu, Lewei Lu, Bin Li, Ping Luo, Tong Lu, Yu~Qiao, and Jifeng Dai. 2023{\natexlab{b}}.
\newblock Internvl: Scaling up vision foundation models and aligning for generic visual-linguistic tasks.
\newblock \emph{arXiv preprint arXiv:2312.14238}.

\bibitem[{Dai et~al.(2023)Dai, Li, Li, Tiong, Zhao, Wang, Li, Fung, and Hoi}]{instructblip}
Wenliang Dai, Junnan Li, Dongxu Li, Anthony Meng~Huat Tiong, Junqi Zhao, Weisheng Wang, Boyang Li, Pascale Fung, and Steven Hoi. 2023.
\newblock Instructblip: Towards general-purpose vision-language models with instruction tuning.
\newblock \emph{arXiv preprint arXiv:2305.06500}.

\bibitem[{Dong et~al.(2024)Dong, Zhang, Zang, Cao, Wang, Ouyang, Wei, Zhang, Duan, Cao, Zhang, Li, Yan, Gao, Zhang, Li, Li, Chen, He, Zhang, Qiao, Lin, and Wang}]{internlmxcomposer2}
Xiaoyi Dong, Pan Zhang, Yuhang Zang, Yuhang Cao, Bin Wang, Linke Ouyang, Xilin Wei, Songyang Zhang, Haodong Duan, Maosong Cao, Wenwei Zhang, Yining Li, Hang Yan, Yang Gao, Xinyue Zhang, Wei Li, Jingwen Li, Kai Chen, Conghui He, Xingcheng Zhang, Yu~Qiao, Dahua Lin, and Jiaqi Wang. 2024.
\newblock Internlm-xcomposer2: Mastering free-form text-image composition and comprehension in vision-language large model.
\newblock \emph{arXiv preprint arXiv:2401.16420}.

\bibitem[{Du et~al.(2022)Du, Qian, Liu, Ding, Qiu, Yang, and Tang}]{du2022glm}
Zhengxiao Du, Yujie Qian, Xiao Liu, Ming Ding, Jiezhong Qiu, Zhilin Yang, and Jie Tang. 2022.
\newblock Glm: General language model pretraining with autoregressive blank infilling.
\newblock In \emph{Proceedings of the 60th Annual Meeting of the Association for Computational Linguistics (Volume 1: Long Papers)}, pages 320--335.

\bibitem[{Duncum(2004)}]{duncum2004visual}
Paul Duncum. 2004.
\newblock Visual culture isn't just visual: Multiliteracy, multimodality and meaning.
\newblock \emph{Studies in art education}, 45(3):252--264.

\bibitem[{Fu et~al.(2024)Fu, Chen, Shen, Qin, Zhang, Lin, Yang, Zheng, Li, Sun, Wu, and Ji}]{fu2024mme}
Chaoyou Fu, Peixian Chen, Yunhang Shen, Yulei Qin, Mengdan Zhang, Xu~Lin, Jinrui Yang, Xiawu Zheng, Ke~Li, Xing Sun, Yunsheng Wu, and Rongrong Ji. 2024.
\newblock Mme: A comprehensive evaluation benchmark for multimodal large language models.
\newblock \emph{arXiv preprint arXiv:2306.13394}.

\bibitem[{Gao et~al.(2024)Gao, Chen, Cui, Ren, Wang, Zhu, Tian, Ye, He, Zhu et~al.}]{gao2024mini}
Zhangwei Gao, Zhe Chen, Erfei Cui, Yiming Ren, Weiyun Wang, Jinguo Zhu, Hao Tian, Shenglong Ye, Junjun He, Xizhou Zhu, et~al. 2024.
\newblock Mini-internvl: A flexible-transfer pocket multimodal model with 5\% parameters and 90\% performance.
\newblock \emph{arXiv preprint arXiv:2410.16261}.

\bibitem[{Ge et~al.(2024)Ge, Chen, Chen, Chen, Chen, Chen, Xie, Yan, Zhu, Lin, Dingjie, Wang, Gao, Zhiyi, Li, Wan, and Wang}]{ge2024mllmbenchevaluatingmultimodalllms}
Wentao Ge, Shunian Chen, Guiming~Hardy Chen, Junying Chen, Zhihong Chen, Nuo Chen, Wenya Xie, Shuo Yan, Chenghao Zhu, Ziyue Lin, Song Dingjie, Xidong Wang, Anningzhe Gao, Zhang Zhiyi, Jianquan Li, Xiang Wan, and Benyou Wang. 2024.
\newblock Mllm-bench: Evaluating multimodal llms with per-sample criteria.
\newblock \emph{arXiv preprint arXiv:2311.13951}.

\bibitem[{Goyal et~al.(2017)Goyal, Khot, Summers-Stay, Batra, and Parikh}]{goyal2017making}
Yash Goyal, Tejas Khot, Douglas Summers-Stay, Dhruv Batra, and Devi Parikh. 2017.
\newblock Making the v in vqa matter: Elevating the role of image understanding in visual question answering.

\bibitem[{Hiippala et~al.(2021)Hiippala, Alikhani, Haverinen, Kalliokoski, Logacheva, Orekhova, Tuomainen, Stone, and Bateman}]{hiippala2021ai2d}
Tuomo Hiippala, Malihe Alikhani, Jonas Haverinen, Timo Kalliokoski, Evanfiya Logacheva, Serafina Orekhova, Aino Tuomainen, Matthew Stone, and John~A Bateman. 2021.
\newblock Ai2d-rst: A multimodal corpus of 1000 primary school science diagrams.
\newblock \emph{Language Resources and Evaluation}.

\bibitem[{Hu et~al.(2023)Hu, Yao, Wang, Wang, Pan, Chen, Yu, Wu, Zhao, Zhang, Han, Lin, Xue, Li, Liu, and Sun}]{viscpm}
Jinyi Hu, Yuan Yao, Chongyi Wang, Shan Wang, Yinxu Pan, Qianyu Chen, Tianyu Yu, Hanghao Wu, Yue Zhao, Haoye Zhang, Xu~Han, Yankai Lin, Jiao Xue, Dahai Li, Zhiyuan Liu, and Maosong Sun. 2023.
\newblock Large multilingual models pivot zero-shot multimodal learning across languages.
\newblock \emph{arXiv preprint arXiv:2308.12038}.

\bibitem[{Huang et~al.(1997)Huang, Chen, Chang, and Chen}]{huang1997segmentation}
Chu-Ren Huang, Keh-Jiann Chen, Li-Li Chang, and Feng-Yi Chen. 1997.
\newblock Segmentation standard for chinese natural language processing.
\newblock In \emph{International Journal of Computational Linguistics \& Chinese Language Processing, Volume 2, Number 2, August 1997}, pages 47--62.

\bibitem[{Hwang and Shwartz(2023)}]{hwang2023memecap}
EunJeong Hwang and Vered Shwartz. 2023.
\newblock Memecap: A dataset for captioning and interpreting memes.
\newblock \emph{arXiv preprint arXiv:2305.13703}.

\bibitem[{Kazemzadeh et~al.(2014)Kazemzadeh, Ordonez, Matten, and Berg}]{kazemzadeh2014referitgame}
Sahar Kazemzadeh, Vicente Ordonez, Mark Matten, and Tamara Berg. 2014.
\newblock Referitgame: Referring to objects in photographs of natural scenes.
\newblock In \emph{Proc. of Empirical Methods in Natural Language Processing}.

\bibitem[{Kim et~al.(2023)Kim, Shin, Cho, Jang, Longpre, Lee, Yun, Shin, Kim, Thorne et~al.}]{kim2023prometheus}
Seungone Kim, Jamin Shin, Yejin Cho, Joel Jang, Shayne Longpre, Hwaran Lee, Sangdoo Yun, Seongjin Shin, Sungdong Kim, James Thorne, et~al. 2023.
\newblock Prometheus: Inducing fine-grained evaluation capability in language models.
\newblock \emph{Proc. of International Conference on Learning Representations}.

\bibitem[{Kim et~al.(2024)Kim, Suk, Longpre, Lin, Shin, Welleck, Neubig, Lee, Lee, and Seo}]{kim2024prometheus2opensource}
Seungone Kim, Juyoung Suk, Shayne Longpre, Bill~Yuchen Lin, Jamin Shin, Sean Welleck, Graham Neubig, Moontae Lee, Kyungjae Lee, and Minjoon Seo. 2024.
\newblock Prometheus 2: An open source language model specialized in evaluating other language models.
\newblock \emph{arXiv preprint arXiv:2405.01535}.

\bibitem[{Li et~al.(2023)Li, Yang, Liu, Ma, Zhang, Yang, Sun, Liu, and Bai}]{li2023monkey}
Zhang Li, Biao Yang, Qiang Liu, Zhiyin Ma, Shuo Zhang, Jingxu Yang, Yabo Sun, Yuliang Liu, and Xiang Bai. 2023.
\newblock Monkey: Image resolution and text label are important things for large multi-modal models.
\newblock \emph{arXiv preprint arXiv:2311.06607}.

\bibitem[{Lin et~al.(2014)Lin, Maire, Belongie, Hays, Perona, Ramanan, Doll{\'a}r, and Zitnick}]{lin2014microsoft}
Tsung-Yi Lin, Michael Maire, Serge Belongie, James Hays, Pietro Perona, Deva Ramanan, Piotr Doll{\'a}r, and C~Lawrence Zitnick. 2014.
\newblock Microsoft coco: Common objects in context.
\newblock In \emph{Proc. of European Conference on Computer Vision}.

\bibitem[{Liu et~al.(2024)Liu, Li, Wu, and Lee}]{liu2024visual}
Haotian Liu, Chunyuan Li, Qingyang Wu, and Yong~Jae Lee. 2024.
\newblock Visual instruction tuning.
\newblock \emph{Proc. of Neural Information Processing Systems}.

\bibitem[{Liu et~al.(2023{\natexlab{a}})Liu, Duan, Zhang, Li, Zhang, Zhao, Yuan, Wang, He, Liu et~al.}]{liu2023mmbench}
Yuan Liu, Haodong Duan, Yuanhan Zhang, Bo~Li, Songyang Zhang, Wangbo Zhao, Yike Yuan, Jiaqi Wang, Conghui He, Ziwei Liu, et~al. 2023{\natexlab{a}}.
\newblock Mmbench: Is your multi-modal model an all-around player?
\newblock \emph{arXiv preprint arXiv:2307.06281}.

\bibitem[{Liu et~al.(2023{\natexlab{b}})Liu, Li, Li, Yu, Huang, Peng, Liu, Chen, Li, Jin et~al.}]{liu2023hidden}
Yuliang Liu, Zhang Li, Hongliang Li, Wenwen Yu, Mingxin Huang, Dezhi Peng, Mingyu Liu, Mingrui Chen, Chunyuan Li, Lianwen Jin, et~al. 2023{\natexlab{b}}.
\newblock On the hidden mystery of ocr in large multimodal models.
\newblock \emph{arXiv preprint arXiv:2305.07895}.

\bibitem[{Lu et~al.(2024{\natexlab{a}})Lu, Liu, Zhang, Wang, Dong, Liu, Sun, Ren, Li, Yang, Sun, Deng, Xu, Xie, and Ruan}]{lu2024deepseekvl}
Haoyu Lu, Wen Liu, Bo~Zhang, Bingxuan Wang, Kai Dong, Bo~Liu, Jingxiang Sun, Tongzheng Ren, Zhuoshu Li, Hao Yang, Yaofeng Sun, Chengqi Deng, Hanwei Xu, Zhenda Xie, and Chong Ruan. 2024{\natexlab{a}}.
\newblock Deepseek-vl: Towards real-world vision-language understanding.
\newblock \emph{arXiv preprint arXiv:2403.05525}.

\bibitem[{Lu et~al.(2024{\natexlab{b}})Lu, Bansal, Xia, Liu, Li, Hajishirzi, Cheng, Chang, Galley, and Gao}]{lu2023mathvista}
Pan Lu, Hritik Bansal, Tony Xia, Jiacheng Liu, Chunyuan Li, Hannaneh Hajishirzi, Hao Cheng, Kai-Wei Chang, Michel Galley, and Jianfeng Gao. 2024{\natexlab{b}}.
\newblock Mathvista: Evaluating mathematical reasoning of foundation models in visual contexts.
\newblock \emph{Proc. of International Conference on Learning Representations}.

\bibitem[{Marino et~al.(2019)Marino, Rastegari, Farhadi, and Mottaghi}]{marino2019okvqa}
Kenneth Marino, Mohammad Rastegari, Ali Farhadi, and Roozbeh Mottaghi. 2019.
\newblock Ok-vqa: A visual question answering benchmark requiring external knowledge.
\newblock In \emph{Proc. of Computer Vision and Pattern Recognition}.

\bibitem[{Masry et~al.(2022)Masry, Long, Tan, Joty, and Hoque}]{masry2022chartqa}
Ahmed Masry, Do~Xuan Long, Jia~Qing Tan, Shafiq Joty, and Enamul Hoque. 2022.
\newblock Chartqa: A benchmark for question answering about charts with visual and logical reasoning.
\newblock \emph{arXiv preprint arXiv:2203.10244}.

\bibitem[{Mathew et~al.(2021)Mathew, Karatzas, and Jawahar}]{mathew2021docvqa}
Minesh Mathew, Dimosthenis Karatzas, and CV~Jawahar. 2021.
\newblock Docvqa: A dataset for vqa on document images.

\bibitem[{OpenAI.(2024)}]{gpt4o}
OpenAI. 2024.
\newblock Gpt4o.
\newblock \emph{https://openai.com/index/hello-gpt-4o}.

\bibitem[{Plummer et~al.(2015)Plummer, Wang, Cervantes, Caicedo, Hockenmaier, and Lazebnik}]{plummer2015flickr30k}
Bryan~A Plummer, Liwei Wang, Chris~M Cervantes, Juan~C Caicedo, Julia Hockenmaier, and Svetlana Lazebnik. 2015.
\newblock Flickr30k entities: Collecting region-to-phrase correspondences for richer image-to-sentence models.
\newblock In \emph{Proc. of International Conference on Computer Vision}.

\bibitem[{Rasley et~al.(2020)Rasley, Rajbhandari, Ruwase, and He}]{rasley2020deepspeed}
Jeff Rasley, Samyam Rajbhandari, Olatunji Ruwase, and Yuxiong He. 2020.
\newblock Deepspeed: System optimizations enable training deep learning models with over 100 billion parameters.
\newblock In \emph{Proceedings of the 26th ACM SIGKDD International Conference on Knowledge Discovery \& Data Mining}, pages 3505--3506.

\bibitem[{Schwenk et~al.(2022)Schwenk, Khandelwal, Clark, Marino, and Mottaghi}]{schwenk2022okvqa}
Dustin Schwenk, Apoorv Khandelwal, Christopher Clark, Kenneth Marino, and Roozbeh Mottaghi. 2022.
\newblock A-okvqa: A benchmark for visual question answering using world knowledge.

\bibitem[{Singh et~al.(2019)Singh, Natarajan, Shah, Jiang, Chen, Batra, Parikh, and Rohrbach}]{singh2019towards}
Amanpreet Singh, Vivek Natarajan, Meet Shah, Yu~Jiang, Xinlei Chen, Dhruv Batra, Devi Parikh, and Marcus Rohrbach. 2019.
\newblock Towards vqa models that can read.

\bibitem[{Singh et~al.(2021)Singh, Pang, Toh, Huang, Galuba, and Hassner}]{singh2021textocr}
Amanpreet Singh, Guan Pang, Mandy Toh, Jing Huang, Wojciech Galuba, and Tal Hassner. 2021.
\newblock Textocr: Towards large-scale end-to-end reasoning for arbitrary-shaped scene text.

\bibitem[{Wang et~al.(2024)Wang, Bai, Tan, Wang, Fan, Bai, Chen, Liu, Wang, Ge, Fan, Dang, Du, Ren, Men, Liu, Zhou, Zhou, and Lin}]{Qwen2VL}
Peng Wang, Shuai Bai, Sinan Tan, Shijie Wang, Zhihao Fan, Jinze Bai, Keqin Chen, Xuejing Liu, Jialin Wang, Wenbin Ge, Yang Fan, Kai Dang, Mengfei Du, Xuancheng Ren, Rui Men, Dayiheng Liu, Chang Zhou, Jingren Zhou, and Junyang Lin. 2024.
\newblock Qwen2-vl: Enhancing vision-language model's perception of the world at any resolution.
\newblock \emph{arXiv preprint arXiv:2409.12191}.

\bibitem[{Wang et~al.(2023)Wang, Lv, Yu, Hong, Qi, Wang, Ji, Yang, Zhao, Song et~al.}]{wang2023cogvlm}
Weihan Wang, Qingsong Lv, Wenmeng Yu, Wenyi Hong, Ji~Qi, Yan Wang, Junhui Ji, Zhuoyi Yang, Lei Zhao, Xixuan Song, et~al. 2023.
\newblock Cogvlm: Visual expert for pretrained language models.
\newblock \emph{arXiv preprint arXiv:2311.03079}.

\bibitem[{Xu et~al.(2024)Xu, Yao, Guo, Cui, Ni, Ge, Chua, Liu, and Huang}]{xu2024llava-uhd}
Ruyi Xu, Yuan Yao, Zonghao Guo, Junbo Cui, Zanlin Ni, Chunjiang Ge, Tat-Seng Chua, Zhiyuan Liu, and Gao Huang. 2024.
\newblock {LLaVA-UHD}: an lmm perceiving any aspect ratio and high-resolution images.
\newblock \emph{arXiv preprint arXiv:2403.11703}.

\bibitem[{Yu et~al.(2016)Yu, Poirson, Yang, Berg, and Berg}]{yu2016modeling}
Licheng Yu, Patrick Poirson, Shan Yang, Alexander~C Berg, and Tamara~L Berg. 2016.
\newblock Modeling context in referring expressions.
\newblock In \emph{Proc. of European Conference on Computer Vision}.

\bibitem[{Yu et~al.(2023{\natexlab{a}})Yu, Yao, Zhang, He, Han, Cui, Hu, Liu, Zheng, Sun et~al.}]{yu2023rlhf}
Tianyu Yu, Yuan Yao, Haoye Zhang, Taiwen He, Yifeng Han, Ganqu Cui, Jinyi Hu, Zhiyuan Liu, Hai-Tao Zheng, Maosong Sun, et~al. 2023{\natexlab{a}}.
\newblock Rlhf-v: Towards trustworthy mllms via behavior alignment from fine-grained correctional human feedback.
\newblock \emph{arXiv preprint arXiv:2312.00849}.

\bibitem[{Yu et~al.(2024)Yu, Zhang, Yao, Dang, Chen, Lu, Cui, He, Liu, Chua, and Sun}]{yu2024rlaifv}
Tianyu Yu, Haoye Zhang, Yuan Yao, Yunkai Dang, Da~Chen, Xiaoman Lu, Ganqu Cui, Taiwen He, Zhiyuan Liu, Tat-Seng Chua, and Maosong Sun. 2024.
\newblock Rlaif-v: Aligning mllms through open-source ai feedback for super gpt-4v trustworthiness.
\newblock \emph{arXiv preprint arXiv:2405.17220}.

\bibitem[{Yu et~al.(2023{\natexlab{b}})Yu, Yang, Li, Wang, Lin, Liu, Wang, and Wang}]{yu2023mmvet}
Weihao Yu, Zhengyuan Yang, Linjie Li, Jianfeng Wang, Kevin Lin, Zicheng Liu, Xinchao Wang, and Lijuan Wang. 2023{\natexlab{b}}.
\newblock Mm-vet: Evaluating large multimodal models for integrated capabilities.
\newblock \emph{arXiv preprint arXiv:2308.02490}.

\bibitem[{Yue et~al.(2023)Yue, Ni, Zhang, Zheng, Liu, Zhang, Stevens, Jiang, Ren, Sun et~al.}]{yue2023mmmu}
Xiang Yue, Yuansheng Ni, Kai Zhang, Tianyu Zheng, Ruoqi Liu, Ge~Zhang, Samuel Stevens, Dongfu Jiang, Weiming Ren, Yuxuan Sun, et~al. 2023.
\newblock Mmmu: A massive multi-discipline multimodal understanding and reasoning benchmark for expert agi.
\newblock \emph{arXiv preprint arXiv:2311.16502}.

\bibitem[{Zeng et~al.(2022)Zeng, Liu, Du, Wang, Lai, Ding, Yang, Xu, Zheng, Xia et~al.}]{zeng2022glm}
Aohan Zeng, Xiao Liu, Zhengxiao Du, Zihan Wang, Hanyu Lai, Ming Ding, Zhuoyi Yang, Yifan Xu, Wendi Zheng, Xiao Xia, et~al. 2022.
\newblock Glm-130b: An open bilingual pre-trained model.
\newblock \emph{arXiv preprint arXiv:2210.02414}.

\bibitem[{Zhang et~al.(2023)Zhang, Lu, Wang, Yan, Yan, Qin, Wang, Yan, Wang, and Petzold}]{zhang2023gpt4visiongeneralistevaluatorvisionlanguage}
Xinlu Zhang, Yujie Lu, Weizhi Wang, An~Yan, Jun Yan, Lianke Qin, Heng Wang, Xifeng Yan, William~Yang Wang, and Linda~Ruth Petzold. 2023.
\newblock Gpt-4v(ision) as a generalist evaluator for vision-language tasks.
\newblock \emph{arXiv preprint arXiv:2311.01361}.

\bibitem[{Zheng et~al.(2024)Zheng, Chiang, Sheng, Zhuang, Wu, Zhuang, Lin, Li, Li, Xing et~al.}]{zheng2024judging}
Lianmin Zheng, Wei-Lin Chiang, Ying Sheng, Siyuan Zhuang, Zhanghao Wu, Yonghao Zhuang, Zi~Lin, Zhuohan Li, Dacheng Li, Eric Xing, et~al. 2024.
\newblock Judging llm-as-a-judge with mt-bench and chatbot arena.
\newblock \emph{Proc. of Neural Information Processing Systems}.

\end{thebibliography}

\clearpage
\appendix
\section{Appendix}
\label{sec:appendix}




\subsection{Detailed category definition}
\label{benchdetail}

We provide a detailed definition of each task in AlignMMBench below, briefed in Figure~\ref{fig:PieChart}.

\paragraph{Perception and understanding.} This category involves answering questions by synthesizing information from images and world knowledge. It comprises six different tasks:
\begin{itemize}[leftmargin=*, noitemsep, topsep=0pt]
\item \textbf{Description.} This task evaluates the model's ability to provide general descriptions of an image. All questions are constructed by sampling from a list of description requests. 
\item \textbf{Recognition.} This task requires the model to recognize objects in the image, covering a wide range of topics, including plants, animals, famous people, and films.
\item \textbf{Counting.} This task requires the model to count specific objects in the image.
\item \textbf{OCR.} This task assesses the model's ability to understand Chinese characters within images.
\item \textbf{Meme.} Memes consist of widely shared images paired with text, using visual metaphors to convey meanings different from the original images~\cite{hwang2023memecap}. This task aims to test whether models can truly understand the meaning of images, which is useful for examining the model's ability to grasp the profound metaphor between images and texts.
\item \textbf{Knowledge.} This task tests the model's world knowledge. VLMs are required to provide comprehensive answers based on their internal knowledge, such as introducing the history of a tourist attraction or providing public information about a celebrity.
\end{itemize}

\paragraph{Reasoning and analysis.} Beyond perception \& understanding, this section focuses on assessing the model's capabilities in information processing and analysis. These tasks often require numerical calculations or logical reasoning to provide accurate responses. This category is divided into five distinct tasks:

\begin{itemize}[leftmargin=*, noitemsep, topsep=0pt]
\item \textbf{Reasoning.} This task assesses the model’s logical reasoning capacity, requiring it to make accurate choices and interpretations by identifying corresponding patterns in images, as illustrated in Figure \ref{fig:example}.
\item \textbf{Chart.} This task requires the model to parse information from charts in images, converting it to specified formats or calculating question-specific statistical characteristics.
\item \textbf{Problem.} This task involves solving high school-level problems across various disciplines such as Mathematics, Physics, Chemistry, and Biology. To differentiate from the OCR category, we ensure that most of these problems cannot be correctly answered based solely on the text in the images.
\item \textbf{Comparison.} This task requires models to identify differences between two sub-images or compare objects in different locations. Models need to locate objects and analyze the differences between them.
\item \textbf{Writing.} Writing based on images requires comprehensive analysis and creative abilities. In this task, the model is required to compose poems, advertisements, speeches, and other forms of text based on the images.
\end{itemize}

\paragraph{Dialogue context.} In real-world user scenarios, models are typically required to generate responses based on dialogue context. However, this remains an underexplored area in existing multimodal benchmarks. We broadly categorize this capability into two tasks:
\begin{itemize}[leftmargin=*, noitemsep, topsep=0pt]
\item \textbf{Coherence.} In practical applications, users prefer that VLMs retain and utilize the entire dialogue context to generate coherent responses. This task focuses on evaluating the model's ability to follow instructions and incorporate information from dialogue history.
\item \textbf{Incoherence.} When a model provides incorrect or irrelevant responses, users usually highlight these errors and expect the model to correct their replies. Therefore, this task assesses the model's self-correcting abilities.
\end{itemize}





\begin{table*}[h]
\centering
\small
\setlength{\tabcolsep}{0.8mm}{ 
\begin{tabular}{l|l|l|l|l}
\toprule

Model         & Size & Backbone/version                                               & Main Language & Avg Scores \\
\midrule
Qwen2-VL       & 72B  & Qwen2-72B+QwenViT & Chinese       & 6.51       \\
CogVLM2       & 19B  & \makecell[l]{EVA2-CLIP-E + Adapter + MLP + LLaMA3 \\ with vision Expert} & Chinese       & 5.81       \\
InternVL-V1.5 & 26B  & ViT-L/14 + InternLM2                                    & Chinese       & 5.62       \\
InternVL2 & 76B  & InternViT-6B + Llama-3-70B-Instruct                                    & Chinese       & 5.57       \\
MiniCPM-V2.5  & 8B   & SigLIP + resampler + LLaMA3                             & Chinese       & 5.42       \\
Qwen-VL-Chat  & 9B   & Vit-BigG + CrossAttn + Qwen                             & Chinese       & 5.13       \\
XComposer2V   & 7B   & ViT-L/14 + InternLM2-7B                                 & Chinese       & 4.97       \\
DeepSeek-VL   & 7B   & SAM-B \& SigLIP-L + Adapter + DeepSeek                  & Chinese       & 4.70       \\
Monkey-Chat   & 9B   & Vit-BigG + resampler + Qwen                             & Chinese       & 4.70       \\
ShareGPT 4V   & 13B  & ViT-L/14 + MLP + LLaMA2                                 & English       &    4.39        \\
LLava1.5      & 13B  & ViT-L/14 + MLP + LLaMA2                                 & English       & 4.31       \\
Yi-VL         & 34B  & ViT-H/14 + MLP + Yi                                     & Chinese       & 4.25       \\
Phi3 Vision   & 4B   & Vit-L/14 + Phi-3-mini                                   & English       & 4.08       \\
InstructBlip  & 9B   & ViT-g/14 + QFormer + LLaMA-7B                           & English       & 3.31      \\
\bottomrule
\end{tabular}}
\caption{Summary table of     
 different open-source model backbones.}
 \label{summarytable}
\end{table*}

\subsection{Radar chart}

\begin{figure}[!h]
\centering
\includegraphics[width=\linewidth]{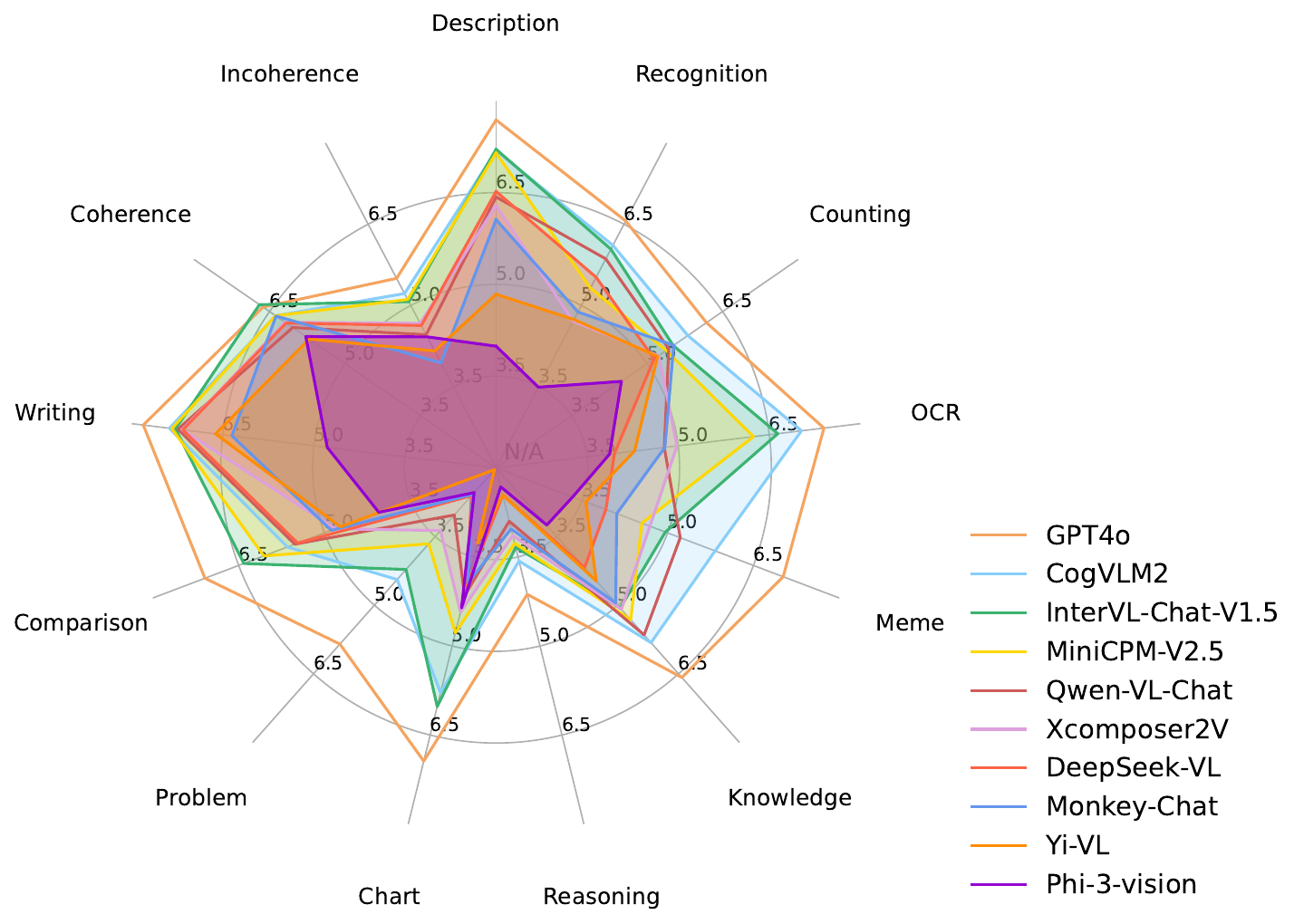}
\caption{Radar chart of leaderboard results.}
\label{fig:radar}
\end{figure}

To better analyze these VLMs' performance on different tasks, we draw an radar chart in Figure~\ref{fig:radar}. Observations from this chart:
\begin{itemize}
    \item GPT4o outperforms other models in most tasks, with CogVLM2 consistently ranking second. Notably, in tasks such as "Problem" and "Meme", GPT4o establishes a significant lead over other VLMs.
    \item Existing VLMs could make good performance in "Perception \& Understanding" category, but still need improvement on "Reasoning \& Analysis" and "Dialogue Context".
\end{itemize}

\begin{figure*}[!h]
\centering
\subfigure[Distribution of questions.]{
\includegraphics[width=0.45\linewidth]{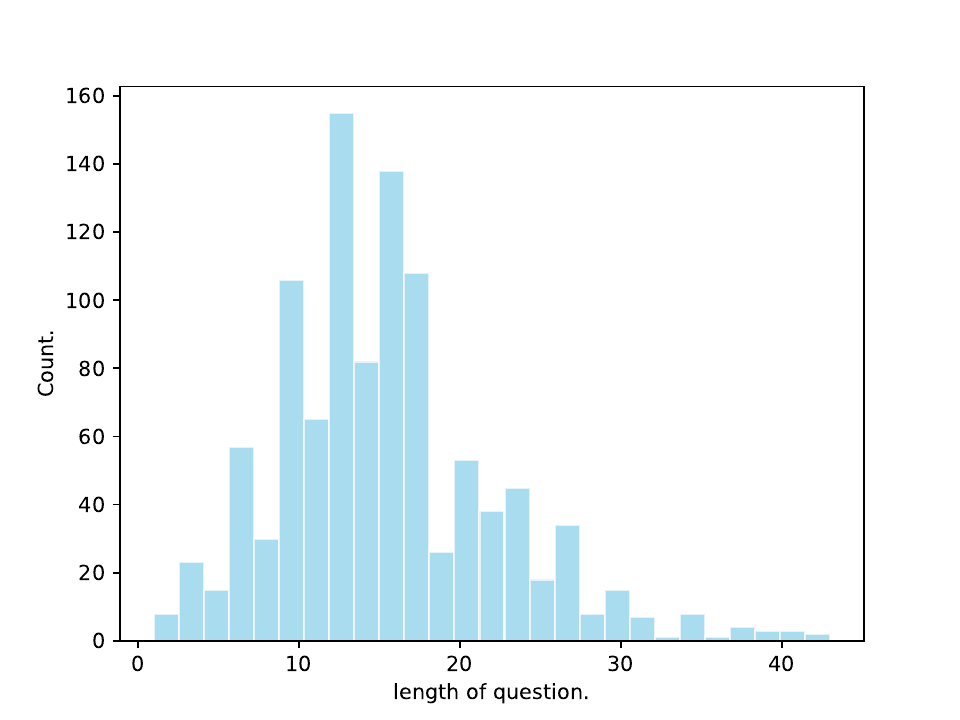}
}
\subfigure[Distribution of reference answers.]{
\includegraphics[width=0.45\linewidth]{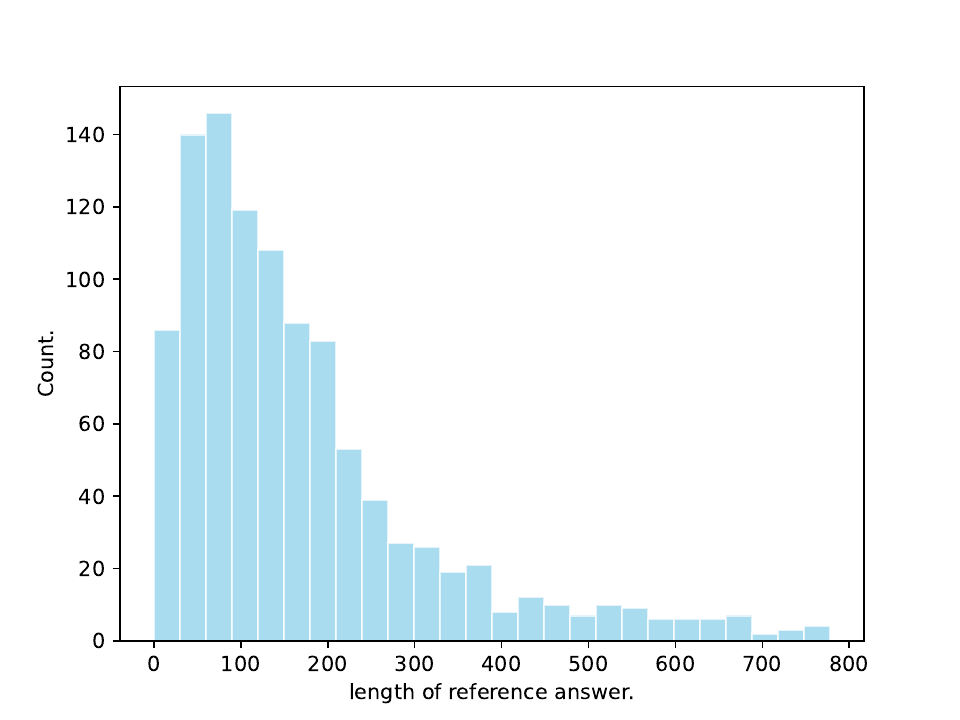}
}
\caption{Length distribution of questions and reference answers.}
\label{fig:dist}
\end{figure*}

\subsection{Prompt rewrite strategy}
\label{augdetail}
In this section, we provide a detailed implementation and example of our prompt rewriting strategy. For test cases in the "Perception \& Understanding" and "Reasoning \& Analysis" categories, we employ GLM4 to generate multiple different prompts while preserving the original intent of the seed prompt. Subsequently, we manually review all prompts to ensure they meet our requirements. The prompt used in our rewriting strategy is illustrated in Figure~\ref{fig:rewriteprompt}.

\begin{figure}[!h]
\centering
\includegraphics[width=\linewidth]{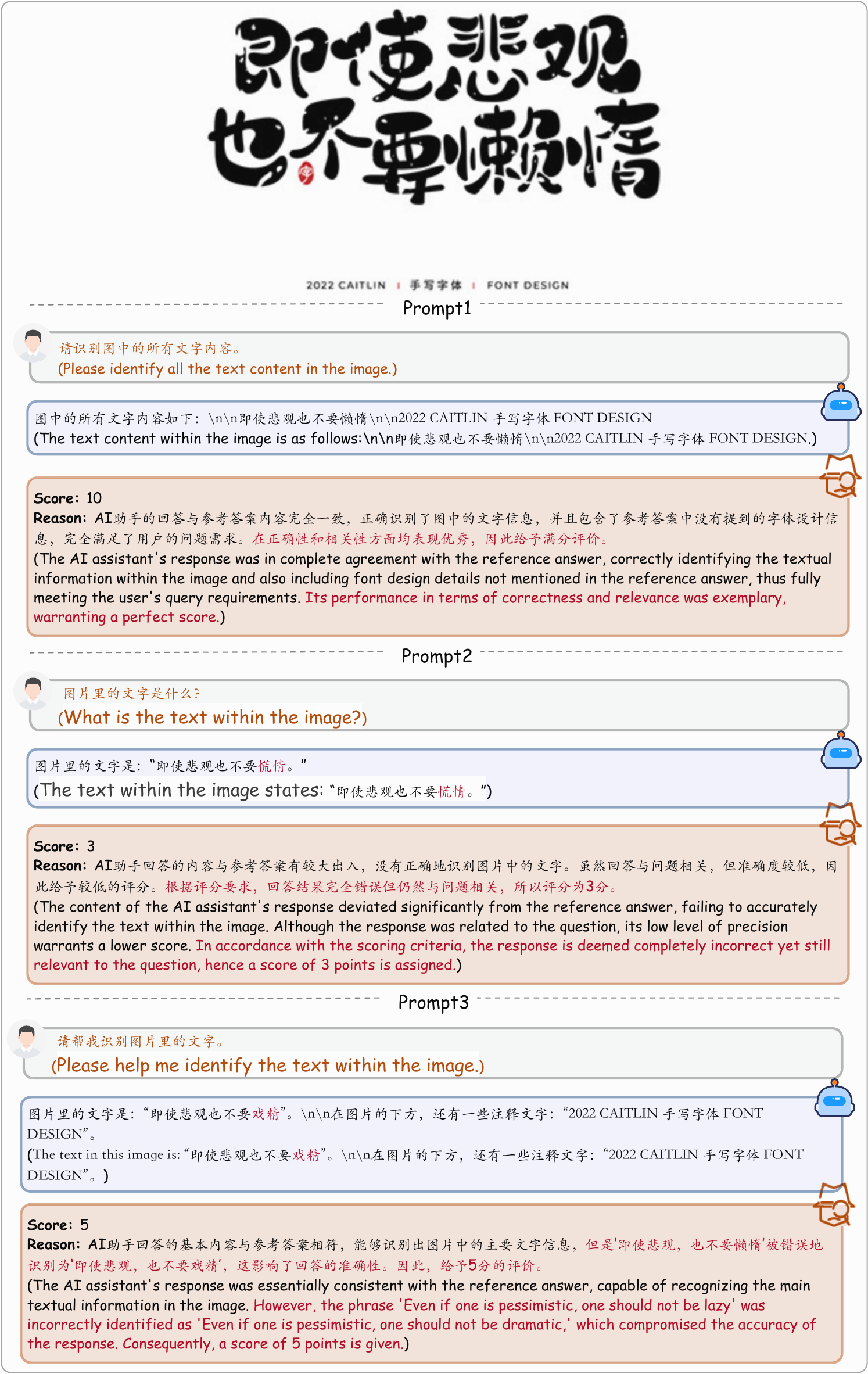}
\caption{An example of equivalent prompts for the same image.}
\label{fig:prompt-rewrite}
\end{figure}

Figure~\ref{fig:prompt-rewrite} illustrates an example of equivalent prompts for the same images, where the model's scores varied significantly. This demonstrates that our rewrite strategy enhances the robustness of our alignment evaluation, effectively minimizing the fluctuations in results attributed to the model's inherent randomness. This improvement ensures more consistent and reliable assessments across different tests.

\subsection{Analysis on question and reference answer lengths}
In this section, we present the distribution of length for questions and reference answers in our AlignMMBench dataset. Results are illustrated in Figure \ref{fig:dist}.

\subsection{More details of data generation}
\subsubsection{Image collection}

We conducted this process in the following steps: 1) Researchers manually defined 13 task types, and construct keywords for each tasks. 2) Web crawlers collected over 20k images using these keywords and removed duplicates based on their hash values. 3) Researchers manually selected images varying in style and content.

Below are several examples of the keywords in step 1:

\begin{itemize}
    \item Description: Poster, Scenery, Flowchart, Map, Cartoon.
    \item Knowledge: Movie, Celebrity, Software icon, Famous attraction, Book.
    \item Chart: Flowchart, Subway map, Bar graph, Table, Index status, Data summary table.
    \item Recognition: Fruit, Flower, Plant, Animal, Food, Simple drawing.
    \item OCR: Handwritten, Documents, Signs, Billboards, Math formulas.
    \item Counting: Fruit stall, Collections, Crowds.
    \item Reasoning: Puzzles, IQ test, Reasoning, Brain teaser.
    \item Problem: Math exercises, Reading comprehension problems, Mechanical problems, Biology problems.
    \item Comparison: Car comparison, Home products comparison, Product comparison, Cosmetics comparison.
    \item Writing: Scenery, Countryside, Forest, Urban, Wildlife.
    \item Meme: Meme.
\end{itemize}

\subsubsection{Human annotation}

\paragraph{Guidelines}

The guidelines provided to human annotators are fundamentally consistent with the prompt detailed in \ref{evalprompt}. Furthermore, annotators were supplied with typical characteristics and illustrative examples corresponding to different scoring levels.

\paragraph{Participants and workload}

The annotators participating in out work are either employees or interns at Zhipu AI. Prior to initiating the annotation process for this specific project, the annotators participated in an initial trial phase. During this phase, they conducted comprehensive discussions with the project leaders to meticulously refine the annotation guidelines and ensure a comprehensive understanding of the task requirements.

The selection of annotators was not limited to individuals from computer science-related disciplines. Although some researchers and a subset of quality inspectors possess significant expertise in computer science, the general annotators were drawn from a variety of academic and professional backgrounds without stringent restrictions or preferences. This diversity serves to mitigate potential educational bias and aids in creating an alignment benchmark that more accurately represents general human preferences.

Approximately ten individuals are involved in the annotation process, and each annotator processes about 60 entries per day. To ensure the high quality of the annotation results, additional personnel perform quality checks on the annotations at the end of each day.

\subsection{Examples}

\subsubsection{Image examples}

\begin{figure}[!h]
\centering
\includegraphics[width=\linewidth]{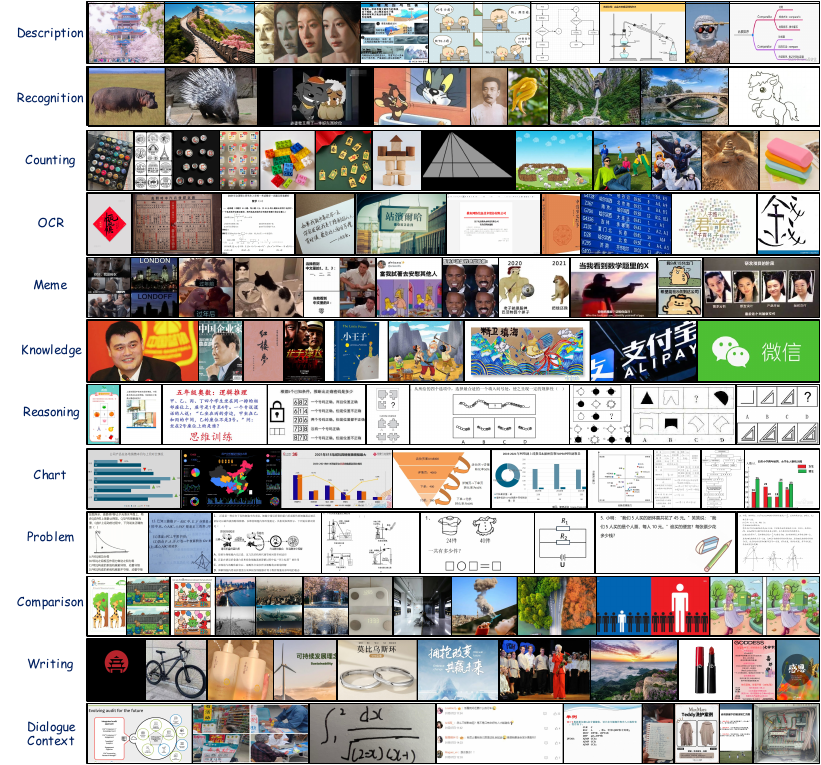}
\caption{Image examples of different tasks.}
\label{fig:images}
\end{figure}

In this section, we randomly selecte multiple images to demonstrate the diversity of visual tasks, as depicted in Figure~\ref{fig:images}. These images illustrate that our AlignMMBench includes a variety of examples that emphasize unique features of the Chinese context. These include: 1) images related to Optical Character Recognition (OCR) and 2) the identification and description of Chinese celebrities, tourist attractions, and holidays. This suggests that our AlignMMBench serves as an excellent supplement to the predominantly English benchmarks currently available.

\begin{CJK*}{UTF8}{gbsn}
\begin{figure}[!h]
\small
\begin{tcolorbox}[colback=red!5!white]
你是一个精通汉语的语言专家。现在我会给你一个用户关于某张图片内容的提问，您需要在不改变提问的整体含义的前提下对该提问进行改写。也就是说，若某个人看到改写前和改写后的提问，他的回答应该是一致的。

[提问内容开始]\{question\}[提问内容结束]

您需要给出 5 次不同的改写结果。输出格式要求输出 5 行，每行包含其中的一个改写结果。请注意不要输出任何其他的字符。

\tcblower
You are a language expert proficient in Chinese. I will now provide you with a user's query about the content of a specific picture. You need to rewrite the query without altering its overall meaning. That is, if someone were to read both the original and the rewritten queries, their responses should be consistent.\\

[Question begin]

\{seed question\}

[Question end] \\

You need to provide five different rewrites of the query. The output format should be five lines, each containing one of the rewrites. Please ensure no other characters are output.
    \end{tcolorbox}
    \caption{The prompt used in prompt-rewrite strategy.}
    \label{fig:rewriteprompt}
\end{figure}
\end{CJK*}

\subsubsection{Evaluation examples}
\begin{figure}[!h]
\centering
\includegraphics[width=\linewidth]{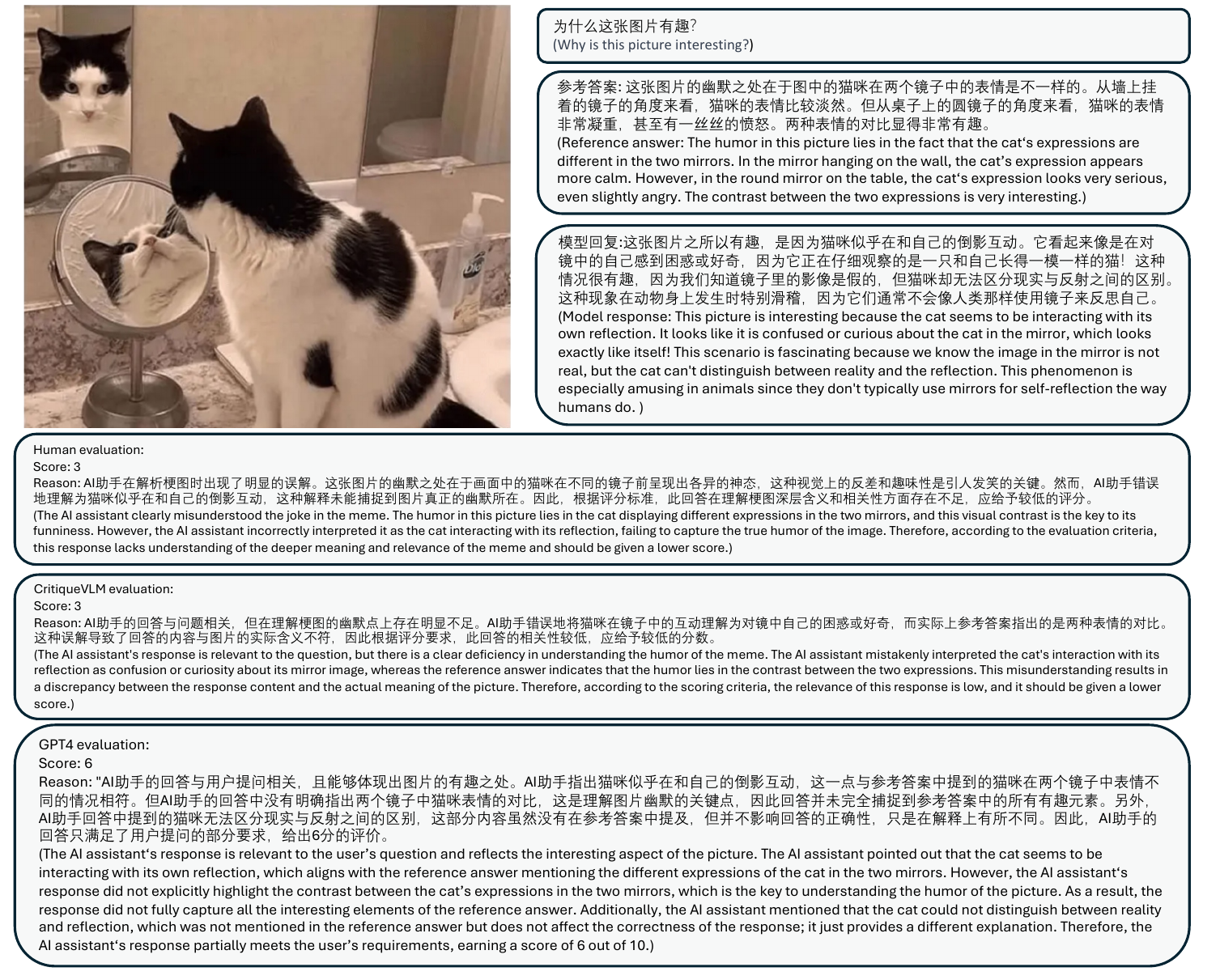}
\caption{A specific comparative example between CritiqueVLM and GPT4 evaluation.}
\label{fig:examples}
\end{figure}

To more effectively illustrate the evaluation capabilities of CritiqueVLM in comparision to GPT-4, we have chosen an evaluation example in Figure ~\ref{fig:examples}.

\subsection{Evaluation prompt}
\label{evalprompt}

Our evaluation prompt consists of general instructions and task-specific prompts, which are listed as follows.

\textbf{General instruction} is illustrated in Figure~\ref{prompt_general}. This prompt encompasses the scoring range, criteria, general hints, and output format. We employ an in-context example to enable the critic model to infer the content of the image from the question and reference answers. Furthermore, it encourages the model to thoughtfully consider details not found in the reference answers rather than immediately deeming them incorrect.

\textbf{Task-specific instruction} is illustrated in Figure~\ref{prompt_spec}. We manually identify the errors made by the API model and construct specific hints to guide it towards accurate evaluations.

\begin{CJK*}{UTF8}{gbsn}
\begin{figure*}[]
\small
\centering
\begin{tcolorbox}
\tiny
\#\# 问答数据

\{对话历史信息\}

- 用户提问： \{问题\}

- 问题类型：\{问题类型\}

[参考答案开始]\{参考答案\}[参考答案结束]

[AI助手回答开始]\{AI助手回答\}[AI助手回答结束]

\#\# 目标

您是一个擅长评价问答质量的专家。请您以客观公正的评判者身份，结合用户提问和参考答案内容，评估AI助手回答的质量。

\#\# 指令

- 评分要求：

1. 您的评估应考虑到正确性、相关性和详细级别等因素。开始您的评估时，请将AI助手回答与参考答案进行比较，并从中找出AI助手回答可能存在的不合理内容，在整体评分结束后提供简短的解释说明。

2. 对于每次评分请求，您需要进行1到10的评分，评分规则如下：整体上，模型回答的质量越高，则分数越高。当AI助手回答完全没有遵循用户的文本指令内容，答非所问、自言自语，总分必须是1分；当AI助手回答和用户提问相关，但内容有害（如：不友好，违反安全性）或语句不通顺不流畅（如：存在大量重复，语言整体混乱），则总分必须是2分；当AI助手回答和用户提问相关，只是回答完全错误，则总分只能是3分。当AI助手回答和用户提问相关，回答有一半以上信息存在事实性错误，则总分应该在4$\sim$ 5分之间。当AI助手回答和用户提问相关，只是有一半以下的信息存在事实性错误，或回答无事实性错误但只满足了用户提问的部分要求，则总分应该是6分。当AI助手回答和用户提问相关，只有极个别错误或回答正确但过于简略，总分应当在7$\sim$ 8分之间。只有当AI助手回答完成了用户提问的所有要求，且提到的所有信息均正确，只是遗漏了一些并不重要的信息，且这些信息不影响回答内容的优劣判断下，总分可以给到9$\sim$ 10分。
    
3. 判断回答与问题是否相关或者回答是否正确时，应该充分考虑语言的同义性，等价性和近似语义。

4. 整体上，在回答结果正确的条件下，有合理分析的回答得分应该更高。

- 注意事项：

由于用户提问属于"\{问题类型\}"类型，因此，在进行打分和解释理由时，应该充分考虑以下注意事项：

\{与类别相关的规则\}

特别地：由于您无法查看图片内容，因此没有提供图片给您。但 AI 助手在回答问题时，是根据图片内容进行回答的。因此当AI助手回答中存在参考答案中没有提到的图片相关信息时，您不能直接认为它是错误的，而是需要结合上述注意事项以及您现有的知识进行综合评价。比如：

    - 问题：这幅卡通图片里的鱼是已经煮熟了吗？我们如何判断？
    
    - 参考答案：是的，这个卡通图中的鱼已经煮熟了，因为它是不透明的，用叉子很容易剥离，并且内部温度已经达到145华氏度。
    
    - AI助手：是的，这幅图片中的鱼已经煮熟了。我们可以通过鱼的颜色、质地和温度来判断。鱼的颜色是粉红色，质地变得松散并且容易用叉子夹碎，温度达到了145°F，这些都是鱼已经煮熟的迹象。
    
    - 评分\&理由：9分。首先，AI助手的结论与参考答案一致。其次，通过颜色和温度判断鱼是否熟了符合生活经验。整体上，AI助手的回答结论正确，逻辑成立，可以给一个较高的分数。
    
\#\# 输出格式

您必须按照以下 JSON 格式输出回答： \{\{ "Rating": , "Reason":\}\}

除了JSON内容外，请不要输出任何其他字符。并且，应该使用中文描述 Reason 部分。

\tcblower
\tiny
\#\# Q\&A Data

\{dialogue history\}

- User Question: \{question\}

- Question Type: \{questiontype\}

[Reference Begin]\{reference answer\}[Reference End]

[AI Assistant Answer Begin]\{AI Assistant Answer\}[AI Assistant Answer End]

\#\# Objects

You are an expert at evaluating the quality of Q\&A responses. As an objective and fair judge, evaluate the AI assistant's response based on the user's question and the reference answer.

\#\# Instructions

- Scoring Requirements:

1. Your evaluation should consider accuracy, relevance, and the level of detail. Begin your assessment by comparing the AI assistant's answer to the reference answer, identify any unreasonable content in the AI assistant's response, and provide a brief explanation after completing the overall score.

2. For each scoring request, you need to score from 1 to 10, with the rules as follows: The higher the quality of the model's response, the higher the score. If the AI assistant completely ignores the user's text instructions, answers unrelated questions or rambles, the total score must be 1; if the AI assistant's response is relevant to the user's question but contains harmful content (such as unfriendly, violating safety) or the sentence is not smooth (such as a lot of repetition, overall language chaos), then the total score must be 2; if the AI assistant's response is relevant to the user's question but completely incorrect, then the total score can only be 3. If the AI assistant's answer is relevant to the user's question and more than half of the information contains factual errors, the total score should be between 4 and 5. If the AI assistant's response is relevant to the user's question, but less than half of the information contains factual errors, or the response has no factual errors but only partially meets the user's question, then the total score should be 6. If the AI assistant's response is relevant to the user's question, has only a few minor errors, or is correct but too brief, the total score should be between 7 and 8. Only when the AI assistant's response meets all the requirements of the user's question, and all the mentioned information is correct, but some unimportant information is omitted and does not affect the judgment of the response quality, the total score can be 9 to 10.

3. When judging whether the answer is relevant to the question or whether the answer is correct, the synonymy, equivalence, and approximate semantics of the language should be fully considered.

4. Overall, if the answer is correct, a well-reasoned response should score higher.

- Points to Note:

Since the user's question falls under the "\{questiontype\}" category, when scoring and explaining reasons, you should consider the following points:

\{Task-Specific Rules\}

Especially: Since you cannot view image content, no images are provided for you. However, the AI assistant's answers are based on the content of the pictures. Therefore, when there is information related to the picture in the AI assistant's answer that is not mentioned in the reference answer, you cannot directly assume it is wrong but need to combine the above points and your existing knowledge for a comprehensive evaluation. For example:

    - Question: Is the fish in this cartoon picture cooked? How can we tell?
    
    - Reference Answer: Yes, the fish in this cartoon picture is cooked because it is opaque, easily peels off with a fork, and the internal temperature has reached 145 degrees Fahrenheit.
    
    - AI Assistant: Yes, the fish in this picture is cooked. We can tell by the color, texture, and temperature of the fish. The fish's color is pink, the texture has become loose and easy to shred with a fork, and the temperature has reached 145°F, which are all signs that the fish is cooked.
    
    - Score \& Reason: 9. Firstly, the AI assistant's conclusion is consistent with the reference answer. Secondly, judging whether the fish is cooked based on its color and temperature aligns with real-life experiences. Overall, the AI assistant's response is correct and logically sound, warranting a high score.
    
\#\# Output Format

You must output the response in the following JSON format: \{\{"Rating": , "Reason":\}\}

Apart from the JSON content, please do not output any other characters. And, the Reason part should be described in Chinese.

\end{tcolorbox}
\caption{The general prompt template for evaluation.}
\label{prompt_general}
\end{figure*}

\begin{figure*}[]
\small
\centering
\begin{tcolorbox}
\tiny

\begin{CJK}{UTF8}{gkai}描述类\end{CJK}: 1. 在评分时，应该充分考虑AI助手回答的组织条理性，逻辑性，语言流畅性和内容完整性。当AI助手回答内容存在不完整时，可以根据不完整程度进行酌情减分，但不能直接认为是AI助手回答是错误的。 2. 由于参考答案是对某张图片内容的完整或者部分描述，因此参考答案可能存在描述不全的情况。当AI助手回答中存在参考答案中不存在的内容时，可以适当怀疑其内容的合理性，但不能直接认为新增内容是错误的。

\begin{CJK}{UTF8}{gkai}推理类\end{CJK}: 1. AI助手回答此类问题时应该提供合理的解释，尤其是问题要求给出理由时。 2. 在评分时，应该首先判断AI助手回答的结论是否正确，若结论错误，可以判定其回答错误，此时应直接给一个低分；若结论正确，再结合其解释的合理性与逻辑性进行综合评分。

\begin{CJK}{UTF8}{gkai}识别类\end{CJK}: 1. 这类问题的回答重点在于识别结果的正确与否，且用户提问和参考答案内容均默认围绕图片进行。您应该耐心地从AI助手回答中找到针对问题的关键答案。当AI助手回答中结果与参考答案语义一致时，务必给高分甚至满分。 2. 即使AI助手的回答和参考答案相比有多余的内容，只要AI助手的识别结果正确，回答中存在和参考答案语义一致的部分，且其余部分符合逻辑，就应当给高分甚至满分。 3. 若回答中对识别结果有合理的描述或者推测，能够酌情加分。当然，不能超过评分规定中的10分。 4. 识别文本内容时，除非题目中特别强调，否则不应该将翻译后的文字视为错误结果。 5. 对于数字，应该注意等价转换，比如 0.1 = 10\%

\begin{CJK}{UTF8}{gkai}计数类\end{CJK}: 1. 这类问题的回答重点在于计数结果的正确与否，且用户提问和参考答案内容均默认围绕图片进行。您应该耐心地从AI助手回答中找到针对问题的关键答案。当AI助手回答中结果与参考答案一致时，务必给高分甚至满分。反之，结果只要不同时，不论差距有多小都必须视为完全错误，必须给低分。 2. 即使AI助手的回答和参考答案相比有很多多余的内容，只要AI助手回答中存在和参考答案语义一致的部分，且其余部分符合逻辑，就应当给高分甚至满分。 3. 若回答中对识别结果有不合理的描述或者推测，应该酌情减分。

\begin{CJK}{UTF8}{gkai}图表类\end{CJK}: 1. 由于您无法查看图片，所以请务必将AI助手回答与参考答案进行比较分析。 2. 对于格式转换题，首先关注回答是否符合新格式要求，其次关注回答的内容是否正确。 3. 对于数字，应该注意等价转换，比如 0.1 = 10\%

\begin{CJK}{UTF8}{gkai}对比类\end{CJK}: 1. 若题目是要求进行对比分析，则有一定组织格式的回答质量优于无组织格式的回答。

\begin{CJK}{UTF8}{gkai}创作类\end{CJK}: 1. 如果提问是要求根据图片写一个故事，那么即便AI助手的回答和参考答案差异较大，也不应该直接给1～4分。而应该根据回答中故事本身的流畅性、戏剧性、有趣程度和与用户提问的关联度等进行打分。

\begin{CJK}{UTF8}{gkai}智力类\end{CJK}: 1. 请注意检查用户提问和AI助手回答的一致性，如果AI助手答非所问，直接给低分。 2. 如果提问类似“这道题怎么做？”，“图片中的题怎么做？”说明用户希望AI助手解决图片中的问题。此时请仔细对比参考答案和AI助手回复来判断AI助手是否真正解决了问题。

\begin{CJK}{UTF8}{gkai}梗图理解\end{CJK}: 1. 这类问题的重点在于评测AI助手能否正确理解该梗图的有趣点。所以你需要仔细比对AI助手和参考答案对图片的解释和理解是否是一种意思，如果是则要打高分； 2. 如果AI助手完全没有解释图片为何有趣，或解释的内容和参考回答相比不足以让用户体会到这张图片背后的真实含义，应该给低分。

\begin{CJK}{UTF8}{gkai}世界知识\end{CJK}: 1. 这类问题的重点在于评测AI助手是否拥有图片以外的世界知识，所以当AI助手回答了一些参考答案中没有的内容时，你不能直接认为它是错误的，而是要结合逻辑是否通顺，与用户提问是否相关，和你自己拥有的知识等方面综合评判。

\begin{CJK}{UTF8}{gkai}OCR理解\end{CJK}: 1. 如果用户提问要求提取或识别内容，则此时应该严格要求答案与参考答案相同，此时不需要考虑“AI助手回答中存在参考答案中没有提到的图片相关信息”，此时只要参考答案与AI助手回答有差距，直接给一个低分。 2. 如果用户提问要求提取或识别内容，此时你只需要检验AI助手回答和参考答案是否匹配，并不需要对回答本身是否包含错误进行检查。 3. 注意判断AI助手回答与参考答案是否语义是一致的，只是语种。若是回答跟参考答案是不同语种的相同表达，则不应该以此理由给1～4分。

\begin{CJK}{UTF8}{gkai}多轮对话\end{CJK}: 1. "对话历史"中呈现了用户和AI助手之间的历史会话内容。 2. 这里问题的重点在于评测AI助手能否在多轮对话中利用之前对话的知识，遵循先前对话的指令。所以您需要深刻理解"对话历史"的每一轮对话，将其与当前的"用户提问"和答案进行对比，给出综合的得分。 3. 如果多轮对话历史中用户要求AI助手修复自己回答的错误，需要仔细观察本次AI助手回复是否能意识到自己的错误并给出正确的答案，如果没有意识到自己的错误或只是承认错误但未给出修正过的答案，则只能获得低分。

\tcblower
\tiny
\textbf{Description}: 1. In grading, the organization, logic, fluency, and completeness of the AI assistant's responses should be fully considered. Points may be deducted for incomplete responses, but incompleteness should not be equated with incorrectness. 2. Since the reference answer may describe an image either partially or completely, it may be incomplete. If the AI assistant includes content not found in the reference answer, its accuracy should be scrutinized, but additional content should not be automatically deemed incorrect.

\textbf{Reasoning}: 1. The AI assistant should provide reasonable explanations, especially when asked to justify a response. 2. When grading, first assess if the conclusion of the AI assistant's response is correct. If incorrect, the response should be considered wrong and receive a low score. If correct, further evaluate the reasonableness and logic of the explanation.

\textbf{Recognition}: 1. These questions focus on the accuracy of identification results, with both user questions and reference answers usually centered around an image. Carefully identify key responses from the AI. If the AI's results align semantically with the reference answer, award high or full marks. 2. Even if the AI’s response includes extra information compared to the reference answer, as long as the identification is correct and parts of the response align semantically with the reference, it deserves high or full marks. 3. Reasonable descriptions or conjectures about identification results in the response can merit additional points, but must not exceed the maximum allowed. 3. When identifying text content, translations should not be considered incorrect unless specifically stated in the question. 5. With numbers, equivalent conversions should be noted, e.g., 0.1 equals 10\%.

\textbf{Counting}: 1. The focus here is on the accuracy of counting results. Patience is required to extract key responses from the AI. If the AI's results match the reference answer, award high or full marks. If not, even slight discrepancies must be treated as completely incorrect, warranting low scores. 2. Even with extraneous content in the AI's response, as long as part of it semantically matches the reference answer and the rest is logical, award high or full marks. 3. Unreasonable descriptions or conjectures about counting results should lead to deductions.

\textbf{Chart}: 1. As you cannot view images, always compare and analyze the AI assistant’s responses against the reference answers. 2. For format conversion questions, prioritize whether the response meets the new format requirements and then check if the response content is correct. 3. For numerical data, ensure equivalent conversions, e.g., 0.1 equals 10\%.

\textbf{Comparison}: 1. If the question involves comparative analysis, organized responses should be rated higher than those without structure.

\textbf{Writing}: 1. If asked to create a story based on an image, significant differences from the reference answer should not automatically result in scores between 1 and 4. Instead, grade based on the narrative flow, drama, interest, and relevance to the query.
Intellectual Type:

\textbf{Problem}: 1. Consistency between the user's query and the AI's response should be verified. Irrelevant answers should receive low scores. 2. For questions like 'How to solve this problem?' or 'How to address the problem shown in the image?', it's crucial to compare the AI's solution to the reference answer to determine if the problem was effectively addressed.

\textbf{Meme}: 1. These questions assess whether the AI understands the humor of a meme. Carefully compare the AI's interpretation with the reference answer. If they align and the AI captures the essence of the humor, award high marks. If the AI fails to explain why the image is humorous or if its explanations fall short of conveying the underlying meaning compared to the reference answer, award low scores.

\textbf{Knowledge}: 1. When the AI provides information not in the reference answer, do not automatically dismiss it as incorrect. Instead, evaluate its relevance to the query, logical coherence, and consistency with known facts.

\textbf{OCR}:  1. When tasked with text extraction or recognition, the response must match the reference answer exactly. If there is a discrepancy, regardless of additional context provided, score low. 2. For text extraction or recognition questions, only assess whether the AI's response matches the reference answer. Do not penalize for potential errors unless the semantic content differs.

\textbf{Dialogue Context}: 1. "Dialogue history" shows the interaction history between the user and the AI. Assess whether the AI uses knowledge from previous exchanges to adhere to the user's ongoing directives. 2. If a user asks the AI to correct an error in its previous responses, observe whether the AI acknowledges and corrects the error in its new response. If it fails to do so, award a low score. 
\end{tcolorbox}
\caption{Task-specific rules in evaluation prompt.}
\label{prompt_spec}
\end{figure*}
\end{CJK*}

\end{document}